\documentclass[aps,twocolumn,floats,pre]{revtex4-1}
\usepackage{graphics,graphicx,epsfig,color}
\usepackage{amssymb,amsmath}
\usepackage{epsf,wrapfig}
\usepackage[mathcal]{euscript}

\usepackage{units} 
\usepackage{commath} 
\usepackage[]{titlesec}
\titleformat{\section}[block]{\color{black}\bfseries\filcenter}{}{1em}{}
\usepackage{hyperref}
\usepackage{upgreek} 
\usepackage{subcaption}
\usepackage{multirow}

\usepackage[english]{babel}

\usepackage[font=small,labelfont=bf,justification=raggedright, format=plain]{caption} 
\titlespacing\section{0pt}{10pt plus 2pt minus 0pt}{10pt plus 2pt minus 2pt}
\titlespacing\subsection{0pt}{5pt plus 2pt minus 0pt}{10pt plus 2pt minus 2pt}

\newcommand{\figtitle}[1]{\textbf{#1}}

\begin{document}

\title{Towards dense object tracking in a 2D honeybee hive}


\title{Towards dense object tracking in a 2D honeybee hive}

\author{Katarzyna Bozek\textsuperscript{\,a},
Laetitia Hebert\textsuperscript{\,a},
Alexander S Mikheyev\textsuperscript{\,a} \&
Greg J Stephens\textsuperscript{\,a,b} 
} 
 \thanks{Corresponding author: kasia.bozek@oist.jp}

\affiliation{
\textsuperscript{a}Okinawa Institute of Science and Technology, 1919-1 Tancha Onna-son, Kunigami-gun, Okinawa 904-0495, Japan\\
 \textsuperscript{b}Department of Physics and Astronomy, VU University Amsterdam, 1081 HV Amsterdam, The Netherlands
}


\begin{abstract}
From human crowds to cells in tissue, the detection and efficient tracking of multiple objects in dense configurations is an important and unsolved problem.  In the past, limitations of image analysis have restricted studies of dense groups to tracking a single or subset of marked individuals, or to coarse-grained group-level dynamics, all of which yield incomplete information.  Here, we combine convolutional neural networks (CNNs) with the model environment of a honeybee hive to automatically recognize  all individuals in a dense group from raw image data.  We create new, adapted individual labeling and use the segmentation architecture U-Net with a loss function dependent on both object identity and orientation. We additionally exploit temporal regularities of the video recording in a recurrent manner and achieve near human-level performance while reducing the network size by $94\%$ compared to the original U-Net architecture. Given our novel application of CNNs, we generate extensive problem-specific image data in which labeled examples are produced through a custom interface with Amazon Mechanical Turk. This dataset contains over 375,000 labeled bee instances  across 720 video frames at $2\,\rm{FPS}$, representing an extensive resource for the development and testing of tracking methods. We correctly detect $96\%$ of individuals with a location error of $\sim7\%$ of a typical body dimension, and orientation error of $12^\circ$, approximating the variability of human raters. Our results provide an important step towards efficient image-based dense object tracking by allowing for the accurate determination of object location and orientation across time-series image data efficiently within one network architecture. 
\end{abstract}
                                                                                                                                                                                                                                                                                                                                  
\keywords{}
\maketitle

\section*{Introduction} 
Image-based dense object tracking is of broad interest in the monitoring of crowd movement as well as the study of collective behavior in biological systems \cite{Li2015-oa}. Automated recognition of individuals in a dense group based on video recording would allow for the efficient implementation of monitoring and tracking frameworks with no additional manual labeling or tracking devices, which are often either impractical or invasive. The challenges in image-based dense object recognition include occlusions and variability in viewpoints and individual appearance. However, recent progress in convolutional neural networks (CNNs) for image segmentation \cite{Long2015-sv}, scene analysis \cite{Pinheiro2014-se}, and object detection \cite{Dai2014-if,He2015-xa,Sermanet2013-ai,Ren2016-pf} represent promising developments towards dense object detection and tracking. Here we apply these tools to a classical unsolved problem in behavioral ecology, the identification of individual organisms in a honeybee hive.

\begin{figure}[t]
  \centering
   \includegraphics[width=0.5\textwidth]{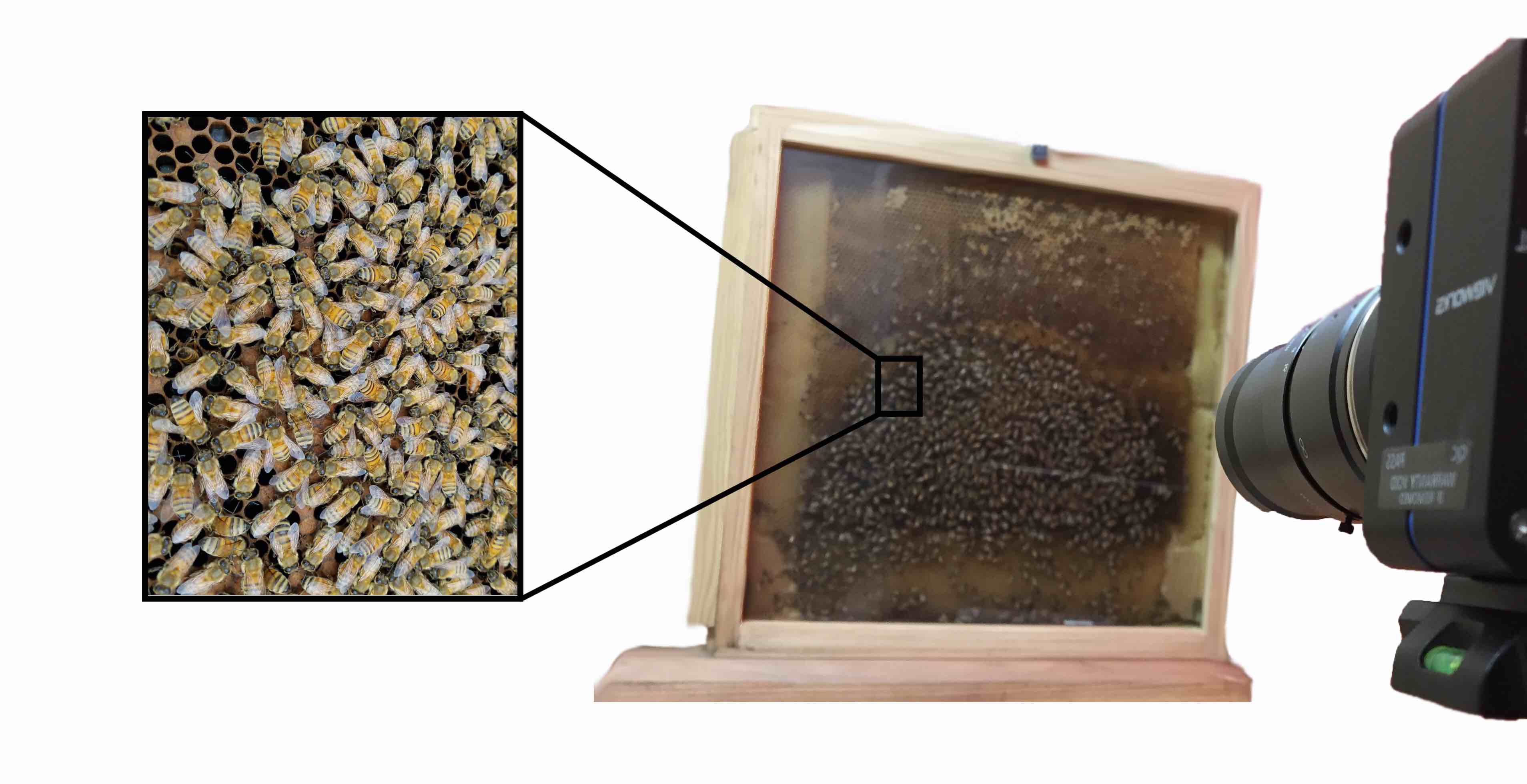}
    \caption{\figtitle{Observation beehive and imaging arrangement.}  Image data was generated from high resolution video recordings of a custom-designed observation beehive in which a honeybee colony was placed on an artificial comb, covered with transparent glass and illuminated with infrared light.  Colonies in the observation hive are approximately two-dimensional and can range in size from hundreds to thousands of individuals.}
    \label{fig:hive}
\end{figure}

Honeybees have long drawn fascination and the study of their behavior has yielded important insights into animal communication, physiology, and neuroscience \cite{Von_Frisch1967-bn,Seeley2010-xn,Winston1991-vd,Karaboga2009-pa}. Honeybees also provide an excellent model system for the study of social behavior as they can be viewed in the natural environment of an  \textit{observation hive} (Fig.~\ref{fig:hive}). However, the complexity of a hive environment presents significant challenges for automated image-based analysis and previous techniques have shown only limited success, particularly under natural conditions \cite{Florea2013-iu,Hendriks2012-en,Kimura2011-oy,Kimura2014-em,Wario2015-fm}. A typical colony consists of hundreds or thousands of closely packed, often occluded, and continually moving individuals. The bees are unevenly distributed over a complex background, the honeycomb, which consists of a variety of different cells containing nectar, pollen, and brood in various stages of development.  If tracking difficulties can be resolved, however, automated image analysis would easily surpass human observers by simultaneously following large numbers of organisms, thus permitting sophisticated studies of social behavior including subtle effects of genetic and molecular perturbations. 

Leveraging high-resolution images of an observation bee hive, we present a method of individual recognition and localization across frames of a video recording. We devise a problem-specific individual labeling, adapt a previously proposed segmentation architecture, and expand its functionality to infer individual bee orientation on the comb. We next strengthen this approach by combining image data in following time frames in a recurrent manner allowing for important reduction of computational cost without compromising the accuracy. As no labeled data for this problem exist, we took advantage of the distributed online marketplace of Amazon Mechanical Turk (AMT) to create extensive training data at modest cost. Our method achieves detection accuracy comparable to human performance on this real-world dense object image data.  Finally, we demonstrate the usefulness of our detection techniques towards a full tracking solution by producing exemplar trajectories with simple registration methods. 

\section*{Related work}
While there have been numerous computer-tracking approaches for the study of social insects, most of them rely on marking individuals, either with simple spots placed on a few individuals \cite{Biesmeijer2005-dr}, or more complex tags with barcodes that distinguish a higher number of individuals \cite{Mersch2013-uf,Wario2015-fm}. Tagging is an obvious solution to recognizing individuals in a dense environment, however, it is laborious, inapplicable to other systems, and impractical on a whole-colony scale. As new individuals emerge, it becomes impossible to mark them without opening and significantly disrupting the colony. Additionally, tag recognition becomes impossible in the situations of partial tag occlusion or viewpoint change \cite{Wario2015-fm}. Due to similar difficulties, previous studies of human crowd tracking were limited to few individuals \cite{Kratz2010-kb,Ali2009-sf} or based on priors about collective motion to aid the performance of tracking algorithms \cite{Ge2012-tz,Rodriguez2011-qo}.

A necessary step towards efficient, image-based dense object tracking is the capacity for individual detection in each frame of a video recording. Recent advances in CNNs have demonstrated their capability to detect and recognize objects in an image (e.g \cite{Girshick2015-tj}). Such object detection methods typically involve an exhaustive sliding window search \cite{Sermanet2013-ai} or a variety of region-based proposals \cite{Hosang2016-lo}. The detection step is then followed by \cite{Sermanet2013-ai} or coupled with \cite{Ren2016-pf,Pinheiro2015-ht} classification of the detected object in the proposed box-shaped region \cite{Sermanet2013-ai,He2015-xa} or a masked patch \cite{Dai2014-if,Pinheiro2015-ht}. Such two-step or two-function architectures were designed for on images containing multi-class, largely variable, and sparse objects.

In contrast, the images of honeybee colonies, cells or human crowds, can contain large numbers of densely packed and highly similar individuals of the same category. In these cases, region-based detection proposals can produce a large list of candidate regions, possibly even covering entire image with distinct objects sharing the same bounding box or mask. Additionally, as each image contains a large number of relatively small objects, keeping the initial image resolution is important for precise object localization. Approximative bounding box estimation \cite{Sermanet2013-ai}, as well as image rescaling \cite{Pinheiro2015-ht} can result in an error margin of the location estimation which is too large for distinguishing among individuals.

Fully convolutional networks \cite{Long2015-sv} allow for image segmentation and categorization on an individual pixel level. These networks are proposal-free and produce label maps for the entire image at its original resolution. Within this framework, each pixel is attributed a category, however, differentiation between instances of objects of the same category is not possible. Instance-aware segmentation has been previously proposed \cite{Dai2015-tz} introducing box-level instance proposals. Images of high-density objects with numerous adjacent individuals necessitate developments allowing for accurate object instance recognition in an efficient manner independent of the number of instances present in the image.

More recently, deep recurrent neural networks (RNNs) were introduced to resolve the task of state estimation with application to the problem of multi-object tracking \cite{Ondruska2016-dd}. Using simulated and real laser sensor measurements this work aimed at predicting the current, unoccluded, complete scene given a sequence of observations capturing only partial information about the scene. A generative probabilistic model inspired by Bayesian filtering \cite{Chen2003-er} was proposed and framed as a RNN architecture allowing for accurate scene estimation even when presented with incomplete observations. The efficacy of this approach however,  was demonstrated entirely on simulated data or simple near-perfect sensor measurements with smooth, linear motion generating black-and-white images where object detection is not part of the tracking task. Here we test the strength of the Bayesian filtering concept on real-world image data comprising dense and cluttered objects with unknown motion dynamics. 

\section{Approach}
\begin{figure}[h]
  \centering
   \includegraphics[width=0.45\textwidth]{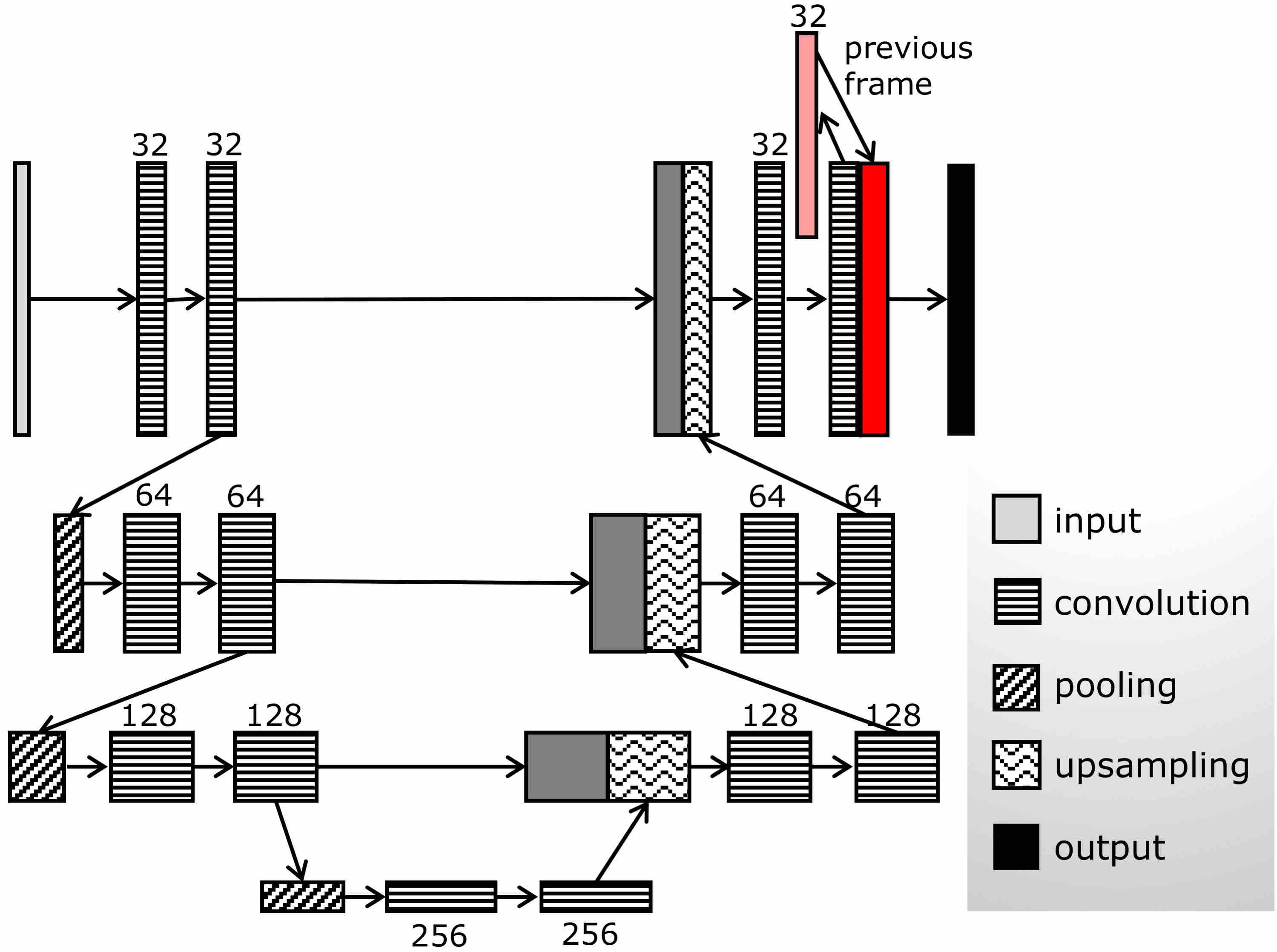}
    \caption{\figtitle{Network architecture.} We used the U-Net architecture with a reduced number of filters and one less pooling and deconvolution steps. A recurrent element was added before the final prediction -- prior representation was stored (pink) and concatenated with the representation of the next image in the time series (red).}\label{fig:unet}
\end{figure}

We propose a solution integrating the fully convolutional neural network U-Net \cite{Ronneberger2015-bx} (Fig.~\ref{fig:unet}) with a recurrent component for accurate object detection in a video sequence. In order to allow for object instance recognition, we defined an adapted labeling covering only the central part of each individual and non-adjacent to other individuals. We demonstrate the capacity of the network to accurately reproduce these labels which additionally allow for recognition of the main axis of each individual. To further indicate the head direction on the main body axis, we propose a loss function approximating individual orientation angle and expand the foreground-background segmentation with object orientation angle estimation. In addition, the recurrent component of the network leverages the information encoded in the video sequence and improves accuracy, while keeping the network at a fraction of the size of the original U-Net. Our proposed approach can localize individuals and recognize their orientation in following frames of a video recording efficiently, in one iteration, without a separate region proposal, sliding window, or masking, thus providing an important foundation for further individual object tracking in a dense group.

\subsection*{Imaging experiment and dataset}
Image data was generated from high-resolution video recordings of a custom-designed observation beehive in which a honeybee colony was placed on one side of a beehive comb, covered with transparent glass and illuminated with infrared light which is imperceptible to the bees (Fig.~\ref{fig:hive})   In brief, the hive was situated on the roof of a laboratory building at OIST graduate university within a prefabricated room of size of 3.6\,m x 2.7\,m x 2.3\,m. The temperature was kept constant at $32.5 ^{\circ}{\rm C}$ and the humidity between 30 and 40\%.  An entrance/exit pipe 20\,cm long connected the hive to the outside. We used a Vieworks Industrial Camera VC series VC-25MX-M72D0-DIN-FM (CMOS sensor, 25 Megapixels, CoaXpress interface, monochrome, F-mount, with image size of 5120 x 5120 pixels) located 1\,m from the hive, so that a typical bee body covered $70\times160$ pixels. The glass surface covered 51\,cm x 51\,cm. Infrared LEDs operating at 850\,nm were mounted around the camera at an angle to avoid reflections. We placed LEDs on four  23\,cm x 22\,cm panels with each panel  equipped with 14 stripes (6 LEDs / strip) for a total of 84 LEDs per panel generating 13.4\,W per panel. Additionally we used one high power infrared spot made of 3 LEDs (ENGIN LZ4-00R608) operating at 850nm and generating 9W. Image data was streamed with four optic fibers to a server where it was compressed without loss and stored with custom software.  The resulting images are in grayscale with 8-bit encoding. The data analyzed here come from two video recordings at $30$\,FPS and $70$\,FPS. For the higher $70$\,FPS time resolution, the infrared light intensity was doubled to compensate for the shorter exposure time. Imaged colonies typically contain greater than 500 individuals.

\subsection*{Data labeling}
We devised a custom javascript interface for manual annotation of bee locations and orientations in the images (Supplemental Fig.~\ref{fig:AMT-schematic}). Through the interface the user defines a bee position and orientation by dragging, dropping, and rotating a bee symbol in an image. An additional round symbol was used to mark the abdomens of bees partially hidden inside of a comb cell where the orientation angle is difficult to determine. We used this interface to generate a labeled image set through AMT. We used 360 frames of the $30$\,FPS and 360 of the $70$\,FPS recording, both down-sampled to $2$\,FPS. In each frame we selected regions of size of $3072\times2048$ and $3072\times3072$ pixels, respectively, containing most of the colony bees against various backgrounds that we submitted for labeling (Supplemental Figs.~\ref{fig:marked_frames30}-\ref{fig:marked_frames70}). As a result we obtained a dataset of $8,640$ $30$\,FPS and $12,960$ $70$\,FPS $512\times512$ pixel images containing total of $375,698$ labeled bees, with an average of $17.4$ bees per image. We also submitted four frames -- two from each recording -- with a total of $2,034$ bee instances for labeling 10-times by independent workers to obtain an estimate of human error in position and angle labeling. This error was calculated as standard deviation of distance of each of the 10 labels to the reference label used in the main dataset for training and testing.

\begin{figure*}[t]
  \centering
    \begin{subfigure}[t]{0.24\textwidth}
        \includegraphics[width=\textwidth]{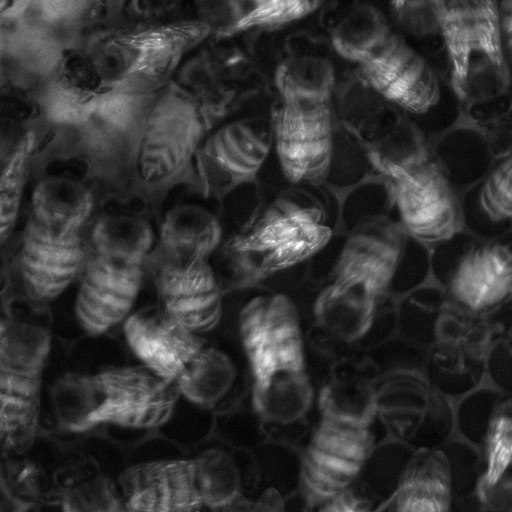}
        \caption{Original image}
    \end{subfigure}
    \begin{subfigure}[t]{0.24\textwidth}
        \includegraphics[width=\textwidth]{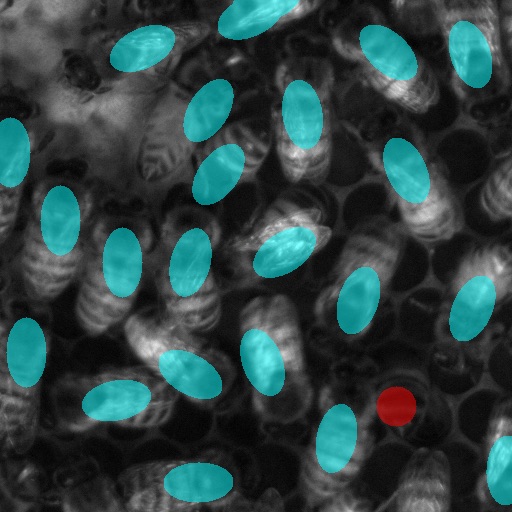}
        \caption{Segmentation labels}
    \end{subfigure}
    \begin{subfigure}[t]{0.24\textwidth}
        \includegraphics[width=\textwidth]{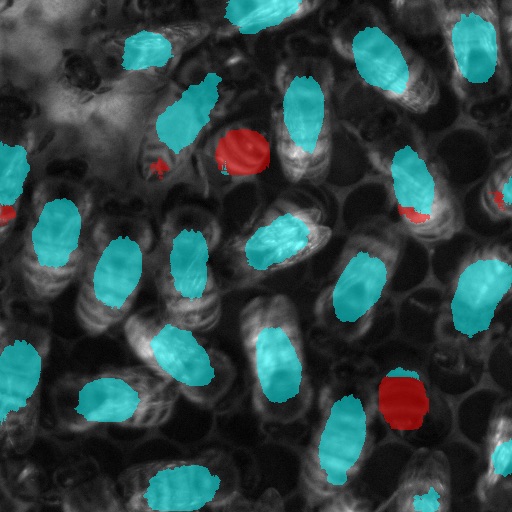}
        \caption{Segmentation results}
    \end{subfigure}
    \begin{subfigure}[t]{0.24\textwidth}
        \includegraphics[width=\textwidth]{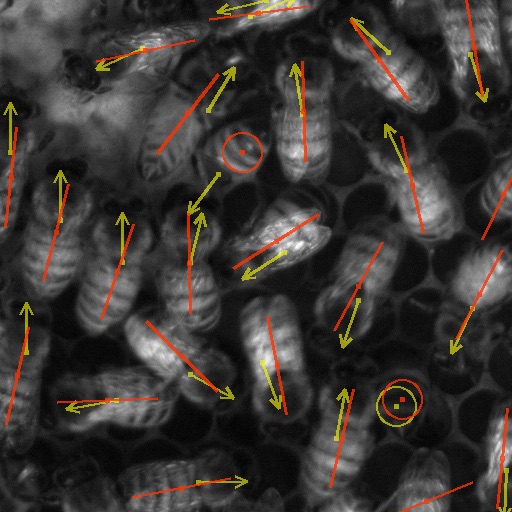}
        \caption{Position and body axis}
    \end{subfigure}
    \caption{\figtitle{Example results for segmentation.} For the original image (a), we show training labels marking bees (b, blue), and abdomens of bees inside honeycomb cells (b, red).  The rest of the image is background. (c) Results of the segmentation network in which each pixel in the input image is classified with a background label, bee label, or abdomen label.  (d) We show predicted locations and body axes estimations (d, red) compared to human labeling (d, yellow). For each contiguously labelled region, the predicted bee location was calculated as the centroid and the predicted body axis was calculated as the angle of the first principal component. Regions representing abdomens are drawn as circles as orientation is ambiguous. Two unlabeled false positives (FP's) are present in this example in the image boundary, as well as a questionable class label mismatch -- a partially visible bee was labeled as fully visible (blue class label) but predicted as bee abdomen (red class label).} \label{fig:segm}
\end{figure*}

\subsection*{Training data}
As the annotation outcome, every individual in an image (e.g. Fig.~\ref{fig:segm}a) was assigned $(x,y,t,\alpha)$ denoting the coordinates of the central point of a bee against the top-left corner of the image, type of the label ($t=1$ when full bee body is visible and $t=2$ when the bee is inside a comb cell), and the body rotation angle $\alpha$ against the vertical pointing upwards and calculated clockwise ($\alpha=0$ if $t=2$). To use this information for segmentation-based individual localization, we generated regions centered over the central point of each bee. For labels with $t=1$ the regions were ellipse-shaped with semi-minor axis $r1=20$ pixels and semi-major axis $r2=35$ pixels, and rotated by the angle $\alpha$ (Fig.~\ref{fig:segm}b). For labels where $t=2$ the regions were circular with $r=20$ pixels. Regions of this shape and size cover the central parts of each bee and are non-adjacent to regions covering neighboring bees in the image. 

To compensate for the class imbalance between foreground bee regions and the non-bee background, we generated weights used for balancing the loss function at every pixel. For every bee region a 2D Gaussian of the same shape was generated, centered over the bee central point, and scaled by either the proportion in the training set of the background pixels to the number of bee-region pixels of the given type $t=1$ and $t=2$ in the task of class segmentation, or scaled by proportion in the training set of the background pixels to the number of bee-region pixels of any type in the task of finding bee orientation angle. 

Training and testing datasets were organized in two ways. First, out of the $21,610$ images $2,176$ were randomly sampled in equal proportions from the $30$\,FPS and $70$\,FPS recording and used as test set. Second, the images were organized in 60 sequences of 360 images of $512\times512$ pixel size. In this time series data the first 324 images of each sequence were used for training and the remaining 36 for testing.

\subsection*{Network and training}
We used the U-Net \cite{Ronneberger2015-bx} segmentation architecture. The number of filters in the initial convolutional layer was doubled after every pooling layer in the expansive path and divided by 2 after each deconvolution in the contracting path (Fig.~\ref{fig:unet}). The convolution kernel size was set as 3.

We first trained the network for foreground-background segmentation with the loss function defined as 3-class softmax scaled by the class imbalance in the entire training set. Next, we expanded the task to finding the direction of each individual orientation. Each foreground pixel, instead of the class label, was set at the value of the bee rotation angle and background pixels were labeled as $-1$. Class identity was not used in this expanded task. The loss function was defined as:
\begin{center}
\(L = w_c \sin(\frac{\alpha - \hat{\alpha}}{2})\),
\end{center}
where \(w_c\) is the class weight and ${\alpha, \hat{\alpha}}$ are the predicted and labeled orientation angle, respectively.

In the network output each contiguous foreground region was interpreted as an individual bee. Foreground patches smaller than $100$ and larger than $6,000$ pixels were discarded, as the label size is $<2,200$ pixels. The centroid location was calculated as the mid-point of all x- and y-coordinates of points in each region. The main body axis was calculated as the angle of the first principal component of the points in each region. In the segmentation task, region class was assigned as the class identity of the majority of pixels within given region. In the bee orientation recognition task, the predicted angle was calculated as the top $0.01$ quantile of all values predicted in the given foreground region. This strategy was motivated by the observation that the orientations in the outer edges of a region are often underestimated, and that the highest value found within a region is closest to the labeled orientation angle. In addition to an independent prediction, the orientation angle was used to assign back and front to the region principal axis.

We additionally expanded the functionality of U-Net to to take advantage of regularities in the image time series patterns. In each pass of the network training or prediction the before-last layer was kept as a prior for the next pass of the network. In the following pass the next image in the time sequence was used as input and the before-last layer was concatenated with the prior representation before calculating network output.

Adaptive moment estimation \cite{Kingma2014-ln} was used during training. Method accuracy was estimated in terms of the capacity to correctly recognize and localize all individuals in an image. We implemented the CNN using Caffe2.

\begin{figure*}[t]
  \centering
    \begin{subfigure}[t]{0.24\textwidth}
        \includegraphics[width=\textwidth]{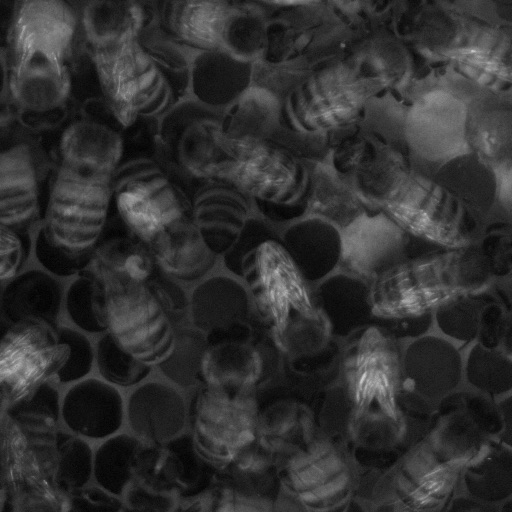}
        \caption{Original image}
    \end{subfigure}
    \begin{subfigure}[t]{0.24\textwidth}
        \includegraphics[width=\textwidth]{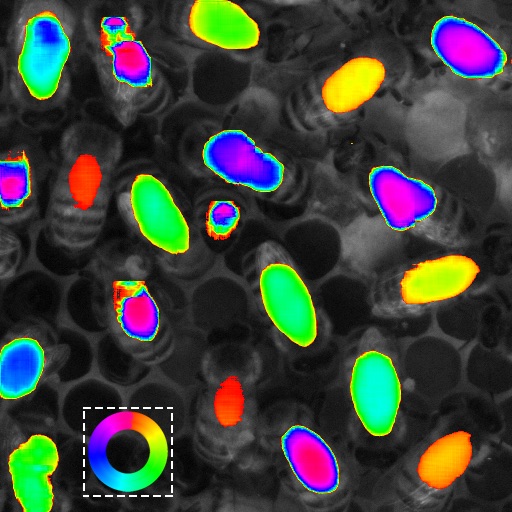}
        \caption{Network output}
    \end{subfigure}
    \begin{subfigure}[t]{0.24\textwidth}
        \includegraphics[width=\textwidth]{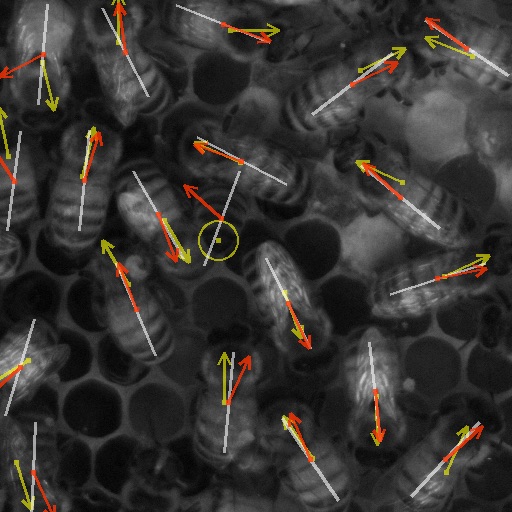}
        \caption{Body axis and angle}
    \end{subfigure}
    \begin{subfigure}[t]{0.24\textwidth}
        \includegraphics[width=\textwidth]{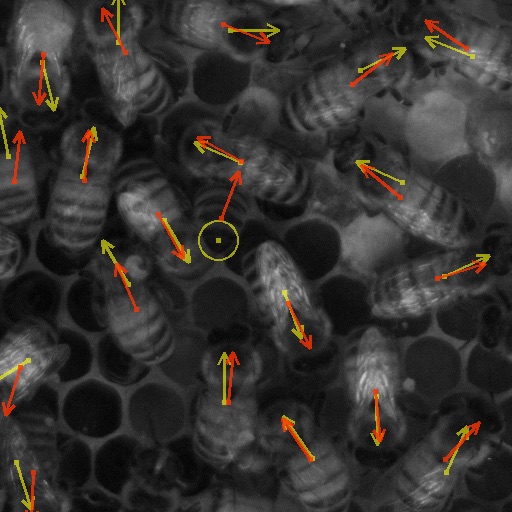}
        \caption{Directed axis}
    \end{subfigure}
    \caption{\figtitle{Example results for body orientation prediction.} For each original image input (a), the network produces orientation predictions (b) for pixels identified as foreground (classes bee or bee abdomen).  Orientation values are represented by the colorwheel within the dashed square.     As in Fig.~3 we estimate body location (c, small squares) and body axis (c, white lines) by computing the centroid and first principal component of contiguous foreground regions, respectively.  The body orientation is separately computed as the mean orientation angle for each region (c, red arrows).  The location and body orientation from human labelling are denoted by yellow arrows. (d) The final predicted body orientation angle is calculated as the body axis with the direction indicated by the estimated angle (d, labels in yellow and predicted directed axis in red). The observation hive is aligned perpendicular to the floor so that a vertically-oriented bee is shown as a vertical arrow.}
     \label{fig:angle}
\end{figure*}

\section*{Results}
\subsection*{Segmentation}
We first tested if individual recognition can be accomplished with the chosen segmentation architecture and two classes of foreground pixels, those that are part of visible bees and those that are part of the abdomens of bees inside the honeycomb cells. We found that the original U-Net architecture resulted in important overfitting and an increase in loss function in the test set (Supplemental Fig.~\ref{fig:dropout}), hence we reduced the size of the U-Net by using 32 filters in the first convolutional layer and eliminating one pooling and one deconvolution layer. This reduced U-Net contained a total of $1.9M$ parameters compared to $31M$ parameters in the original U-Net, thus shrinking the network to just $6\%$ of the original size. Decreasing the number of parameters diminished overfitting. Even so, overfitting was still observable in the reduced network (Supplemental Fig.~\ref{fig:dropout}). We also tested different regularization scenarios using weight decay and dropout \cite{Srivastava_undated-eq}, none of which achieved satisfactory performance within the feasible time span of training (Supplemental Fig.~\ref{fig:dropout}). This could be due to fact that in fully convolutional neural networks, such as U-Net, there is no fully connected layer on which the dropout is usually performed. Different from the fully connected layers, convolutional layers have smaller number of parameters compared to the size of feature maps. Hence, it is believed that convolutional layers suffer less from overfitting and, even though dropout has shown its effectiveness in convolutional layers in some cases \cite{Park_undated-jg,Springenberg2014-xu}, its effect in convolutional layers has not been studied thoroughly.  

We therefore used early-stop in training as a measure against overfitting. In the following, we report network performance after $18$ iterations of training -- the iteration selected based on the increase of loss function of this segmentation network. We apply this stop criterion to training of this segmentation network as well as orientation finding network and recurrent network described below. 

\begin{table*}
\begin{center}
\begin{tabular}{|l|c|c|c|c|c|c|c|}
\hline
& \multirow{3}{*}{TP} & \multirow{3}{*}{FP}   &  \multicolumn{5}{l|}{Error:} \\
\cline{4-8}
 & & & Object & Position & Orientation      & \multirow{2}{*}{Axis [$^\circ$]} & Directed        \\
 & & & class  & [pixel]  & angle [$^\circ$] &  & axis [$^\circ$] \\
\hline\hline
Human labeling        & -    & 0.15 (0.07) & 0.04 & 6.7 & 7.7 & - & - \\
Segmentation          & 0.96 & 0.21 (0.12) & 0.19 & 5.9 & -           & 13.3 (11.2) & - \\
Orientation           & 0.94 & 0.18 (0.10) & -    & 5.6 & 34.0 (32.2) & 15.7 (13.1) & 22.1 (16.7) \\
Recurrent orientation & \textbf{0.96} & \textbf{0.14} (0.06) & - & \textbf{5.1} & \textbf{15.2} (13.3) & \textbf{10.6} (8.8) & \textbf{12.1} (9.7) \\
\hline
\end{tabular}
\end{center}
\caption{\figtitle{Summary results for location and orientation prediction.} In the first row we show the variability among human raters estimated by repeating the labeling task 10 times on an image set. TP-true positives, FP-false positives. As network performance median of error values are listed. Values in brackets are the results after a $50$ pixel margin of the image is discarded, eliminating predictions on partially visible objects. Results cited in the abstract are marked in bold and the full error distributions are presented in Supplemental Fig.~\ref{fig:error_tail}.}
\label{tab:results}
\end{table*}

The segmentation network predicted individual location with a precision of  $\sim6$ pixel on average (Table~\ref{tab:results}), which is similar to the variability among human-assigned labels ($\sim7$ pixels) and much less then a typical bee width of $\sim70$ pixels. While the class prediction was also accurate, there were seemingly high number of false positives (FPs).  However, we noted that most FPs are reported on the image boundary where only incomplete object is visible -- $62\%$ are within $50$ pixel margin of the image (e.g. Fig.~\ref{fig:segm}d, Supplemental Fig.~\ref{fig:unet_hm}). Similarly, the disagreement among human labelers was the highest on the image boundaries, $54\%$ of disagreements are located within $50$ pixel margins of the image, a surface of $< 20\%$ of the size of the images annotated by the raters (Supplemental Fig.~\ref{fig:human_frame}-\ref{fig:boundary_vs_err}). Therefore, in a comprehensive tracking solution, the number of FPs can be reduced by discarding the boundary regions and using overlapping image patches. Additionally, we noticed multiple examples of FPs that, upon a closer inspection were, instead abdomens of bees inside cells that were difficult to spot by human raters (see e.g. Supplemental Fig.~\ref{fig:butts}). Therefore, among the $44\%$ of FPs predicted as bee abdomens, we expect some to be unlabeled true positives (TPs). Foreground class identity -- full bee body vs. abdomen of a partially visible bee -- was incorrectly assigned in $20\%$ of cases however, note that the distinction of the two can often be disputable (e.g. Fig.~\ref{fig:segm}).

We used the elliptical shapes of segmented regions representing bee bodies to deduce the main body axis orientation. In particular, we found that the first principal component of the segmented patches resulted in a relatively precise approximation of each individual orientation with only $13^\circ$ difference on average with the labeled axis orientation (Fig.~\ref{fig:segm}d, Table~\ref{tab:results}).

\subsection*{Location and orientation recognition}
We expanded the segmentation network into an architecture appropriate for the estimation of object orientation angle instead of object category. In this approach, foreground class labels were exchanged with object instance orientation angle. This architecture produced similar performance to the segmentation network with a high TP rate $(0.94)$ and body axis recognition based on the label shape $(16^\circ)$, suggesting that changing the label and loss function did not affect the foreground-background segmentation accuracy.

For the orientation angle, we observed that the error distribution exhibited a small constant baseline component indicative of random predictions (Supplemental Fig.~\ref{fig:error_tail}), and to avoid the undue influence of outlier values  we report median error in our results (Table~\ref{tab:results}). This baseline error can be partially explained by uncertainties at image boundaries, as well as by the variability of angle labeling among human raters (Supplemental Fig.~\ref{fig:human_distribution}-\ref{fig:human_example}), and  is not related to the bee density in the image.  The orientation angle prediction has a median error of $34^\circ$ which is proportionally similar to the axis error given that the head-tail orientation error can range within $[0,180]$ and the axis  within $[0,90]$. Notably, rotation invariance of the CNNs is an unresolved question and more complicated solutions were proposed to address it \cite{Dieleman2015-zi,Marcos2016-es,Worrall2016-up}. It is therefore encouraging that a relatively simple loss function with a reduced U-Net segmentation network allows for approximation of the orientation of the densely packed honeybees. Moreover, the predicted orientation angle can be used merely to indicate the head-tail location on the axis estimated from the shape of the label. In this way we obtained an orientation error of the directed axis $\sim22^\circ$  with this network (Table~\ref{tab:results}).

\subsection*{Recurrent detection and tracking}
We inspected whether regularities in object appearance and movement across time could improve the orientation angle prediction. Image data were organized in a time sequence and, in following iterations of training and testing, consecutive images in the sequence were fed as network input. In each iteration, the penultimate layer of the network was kept as representation of a prior that was concatenated with the same penultimate layer representation of the following image in the sequence in the next iteration of training or testing. In this way network output was a result of both the information in the previous and current time point.

Indeed, we found that incorporating time series image data reduced the error in orientation angle prediction by two-fold $(15^\circ)$ and axis prediction by 2/3 $(11^\circ)$. The orientation error obtained by orienting body axis with the predicted orientation angle was reduced to $12^\circ$ (Fig.~\ref{fig:angle}, Table~\ref{tab:results}), which is significantly better than the non-recurrent approach (Kruskal-Wallis test, $p < 0.0001$) and only marginally higher than the variability observed among human raters. 

\begin{figure}[h]
  \centering
   \includegraphics[width=0.45\textwidth]{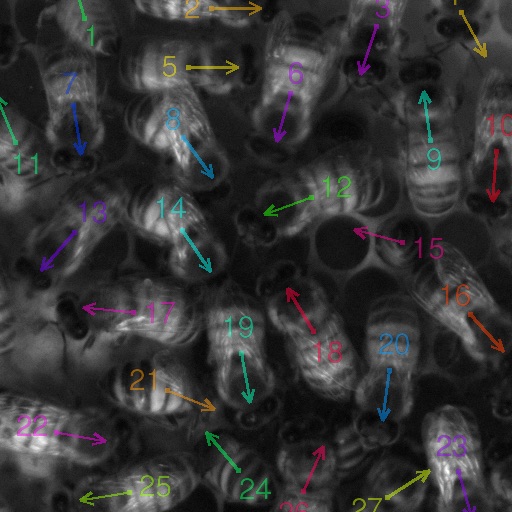}
    \caption{\figtitle{Trajectory reconstruction.} The results of a recurrent approach to object detection allow for trajectory reconstruction using an elementary matching method for registering individuals across frames (see Supplemental Movie).}\label{fig:video}
\end{figure}

Finally, to explore whether our bee detection results could provide the foundation for fully automated image-based tracking, we used elementary ideas to reconstruct bee trajectories.  We matched the closest individuals in following time points and, in case a trajectory is lost, searched up to five frames ahead for a close match that could complete this trajectory. Individual's position, orientation, angle, and velocity were taken into account in the matching. Additionally, short trajectories beginning or ending in the central parts of the image were discarded as potential FPs. As we have no ground truth labels for the individuals' trajectories in our data, we cannot yet quantitatively assess the accuracy of this way performed trajectory estimation. We note however many examples that appear relatively complete (Fig.~\ref{fig:video}, Supplemental Movie) among the 60 sequences of 36 frames of the test set.  

\section{Conclusions}
Accurate individual recognition is an important step towards automated dense object tracking. Here we described an approach for recognizing all individual bees and their orientation in the natural environment of a densely packed honeybee comb. We leverage the power of current segmentation architectures and design labeling to encode additional information about the segmented object -- both in label shape and value -- which allowed us to accurately indicate individual's position and orientation. We additionally enhance our recognition approach through a recursive framework that places the improved accuracy near the level of human labeling, at a strongly reduced computational cost. 

While our principal advance is dense object detection, our results are an important step towards individual trajectory reconstruction as demonstrated with our naive matching approach. Of course, quantitative trajectory reconstruction requires algorithms and analysis beyond the scope of this manuscript. Nevertheless, the positive examples achieved through this simple matching approach, even at low frame-rate ($2$\,FPS), demonstrate that the results provide important steps towards fully automated image-based dense object tracking.

Finally, we suggest that the environment of a honeybee hive offers an excellent model system for the development of tracking approaches. The hive is dense and complex though still tractable for labeling and offers unparalleled access for video recordings. We expect our work to foster significant advances in the quantitative study of this important social organism. In addition, our labeled dataset can be used for the development of other image-based tracking methods and the flexibility of CNN-based segmentation will allow our approach to be usefully applied to a variety of systems.

\subsubsection*{Acknowledgments}
Funding for this work was provided by the OIST Graduate University to ASM and GS. Additional funding was provided by KAKENHI grants 16H06209 and 16KK0175 from the Japan Society for the Promotion of Science to ASM. We are grateful to Yoann Portugal for assistance with colony maintenance and image acquisition, as well as to Quoc-Viet Ha for work on the video acquisition and storage pipeline.

\bibliography{ref}

\begin{thebibliography}{36}%
\makeatletter
\providecommand \@ifxundefined [1]{%
 \@ifx{#1\undefined}
}%
\providecommand \@ifnum [1]{%
 \ifnum #1\expandafter \@firstoftwo
 \else \expandafter \@secondoftwo
 \fi
}%
\providecommand \@ifx [1]{%
 \ifx #1\expandafter \@firstoftwo
 \else \expandafter \@secondoftwo
 \fi
}%
\providecommand \natexlab [1]{#1}%
\providecommand \enquote  [1]{``#1''}%
\providecommand \bibnamefont  [1]{#1}%
\providecommand \bibfnamefont [1]{#1}%
\providecommand \citenamefont [1]{#1}%
\providecommand \href@noop [0]{\@secondoftwo}%
\providecommand \href [0]{\begingroup \@sanitize@url \@href}%
\providecommand \@href[1]{\@@startlink{#1}\@@href}%
\providecommand \@@href[1]{\endgroup#1\@@endlink}%
\providecommand \@sanitize@url [0]{\catcode `\\12\catcode `\$12\catcode
  `\&12\catcode `\#12\catcode `\^12\catcode `\_12\catcode `\%12\relax}%
\providecommand \@@startlink[1]{}%
\providecommand \@@endlink[0]{}%
\providecommand \url  [0]{\begingroup\@sanitize@url \@url }%
\providecommand \@url [1]{\endgroup\@href {#1}{\urlprefix }}%
\providecommand \urlprefix  [0]{URL }%
\providecommand \Eprint [0]{\href }%
\providecommand \doibase [0]{http://dx.doi.org/}%
\providecommand \selectlanguage [0]{\@gobble}%
\providecommand \bibinfo  [0]{\@secondoftwo}%
\providecommand \bibfield  [0]{\@secondoftwo}%
\providecommand \translation [1]{[#1]}%
\providecommand \BibitemOpen [0]{}%
\providecommand \bibitemStop [0]{}%
\providecommand \bibitemNoStop [0]{.\EOS\space}%
\providecommand \EOS [0]{\spacefactor3000\relax}%
\providecommand \BibitemShut  [1]{\csname bibitem#1\endcsname}%
\let\auto@bib@innerbib\@empty
\bibitem [{\citenamefont {Li}\ \emph {et~al.}(2015)\citenamefont {Li},
  \citenamefont {Chang}, \citenamefont {Wang}, \citenamefont {Ni},
  \citenamefont {Hong},\ and\ \citenamefont {Yan}}]{Li2015-oa}%
  \BibitemOpen
  \bibfield  {author} {\bibinfo {author} {\bibfnamefont {T.}~\bibnamefont
  {Li}}, \bibinfo {author} {\bibfnamefont {H.}~\bibnamefont {Chang}}, \bibinfo
  {author} {\bibfnamefont {M.}~\bibnamefont {Wang}}, \bibinfo {author}
  {\bibfnamefont {B.}~\bibnamefont {Ni}}, \bibinfo {author} {\bibfnamefont
  {R.}~\bibnamefont {Hong}}, \ and\ \bibinfo {author} {\bibfnamefont
  {S.}~\bibnamefont {Yan}},\ }\href@noop {} {\  (\bibinfo {year} {2015})},\
  \Eprint {http://arxiv.org/abs/1502.01812} {arXiv:1502.01812 [cs.CV]}
  \BibitemShut {NoStop}%
\bibitem [{\citenamefont {Long}\ \emph {et~al.}(2015)\citenamefont {Long},
  \citenamefont {Shelhamer},\ and\ \citenamefont {Darrell}}]{Long2015-sv}%
  \BibitemOpen
  \bibfield  {author} {\bibinfo {author} {\bibfnamefont {J.}~\bibnamefont
  {Long}}, \bibinfo {author} {\bibfnamefont {E.}~\bibnamefont {Shelhamer}}, \
  and\ \bibinfo {author} {\bibfnamefont {T.}~\bibnamefont {Darrell}},\ }in\
  \href@noop {} {\emph {\bibinfo {booktitle} {Proceedings of the {IEEE}
  Conference on Computer Vision and Pattern Recognition}}}\ (\bibinfo
  {publisher} {cv-foundation.org},\ \bibinfo {year} {2015})\ pp.\ \bibinfo
  {pages} {3431--3440}\BibitemShut {NoStop}%
\bibitem [{\citenamefont {Pinheiro}\ and\ \citenamefont
  {Collobert}(2014)}]{Pinheiro2014-se}%
  \BibitemOpen
  \bibfield  {author} {\bibinfo {author} {\bibfnamefont {P.}~\bibnamefont
  {Pinheiro}}\ and\ \bibinfo {author} {\bibfnamefont {R.}~\bibnamefont
  {Collobert}},\ }in\ \href@noop {} {{\selectlanguage {english}\emph {\bibinfo
  {booktitle} {International Conference on Machine Learning}}}}\ (\bibinfo
  {year} {2014})\ pp.\ \bibinfo {pages} {82--90}\BibitemShut {NoStop}%
\bibitem [{\citenamefont {Dai}\ \emph {et~al.}(2014)\citenamefont {Dai},
  \citenamefont {He},\ and\ \citenamefont {Sun}}]{Dai2014-if}%
  \BibitemOpen
  \bibfield  {author} {\bibinfo {author} {\bibfnamefont {J.}~\bibnamefont
  {Dai}}, \bibinfo {author} {\bibfnamefont {K.}~\bibnamefont {He}}, \ and\
  \bibinfo {author} {\bibfnamefont {J.}~\bibnamefont {Sun}},\ }\href@noop {} {\
   (\bibinfo {year} {2014})},\ \Eprint {http://arxiv.org/abs/1412.1283}
  {arXiv:1412.1283 [cs.CV]} \BibitemShut {NoStop}%
\bibitem [{\citenamefont {He}\ \emph {et~al.}(2015)\citenamefont {He},
  \citenamefont {Zhang}, \citenamefont {Ren},\ and\ \citenamefont
  {Sun}}]{He2015-xa}%
  \BibitemOpen
  \bibfield  {author} {\bibinfo {author} {\bibfnamefont {K.}~\bibnamefont
  {He}}, \bibinfo {author} {\bibfnamefont {X.}~\bibnamefont {Zhang}}, \bibinfo
  {author} {\bibfnamefont {S.}~\bibnamefont {Ren}}, \ and\ \bibinfo {author}
  {\bibfnamefont {J.}~\bibnamefont {Sun}},\ }\href@noop {} {\bibfield
  {journal} {\bibinfo  {journal} {IEEE Trans. Pattern Anal. Mach. Intell.}\
  }\textbf {\bibinfo {volume} {37}},\ \bibinfo {pages} {1904} (\bibinfo {year}
  {2015})}\BibitemShut {NoStop}%
\bibitem [{\citenamefont {Sermanet}\ \emph {et~al.}(2013)\citenamefont
  {Sermanet}, \citenamefont {Eigen}, \citenamefont {Zhang}, \citenamefont
  {Mathieu}, \citenamefont {Fergus},\ and\ \citenamefont
  {LeCun}}]{Sermanet2013-ai}%
  \BibitemOpen
  \bibfield  {author} {\bibinfo {author} {\bibfnamefont {P.}~\bibnamefont
  {Sermanet}}, \bibinfo {author} {\bibfnamefont {D.}~\bibnamefont {Eigen}},
  \bibinfo {author} {\bibfnamefont {X.}~\bibnamefont {Zhang}}, \bibinfo
  {author} {\bibfnamefont {M.}~\bibnamefont {Mathieu}}, \bibinfo {author}
  {\bibfnamefont {R.}~\bibnamefont {Fergus}}, \ and\ \bibinfo {author}
  {\bibfnamefont {Y.}~\bibnamefont {LeCun}},\ }\href@noop {} {\  (\bibinfo
  {year} {2013})},\ \Eprint {http://arxiv.org/abs/1312.6229} {arXiv:1312.6229
  [cs.CV]} \BibitemShut {NoStop}%
\bibitem [{\citenamefont {Ren}\ \emph {et~al.}(2016)\citenamefont {Ren},
  \citenamefont {He}, \citenamefont {Girshick},\ and\ \citenamefont
  {Sun}}]{Ren2016-pf}%
  \BibitemOpen
  \bibfield  {author} {\bibinfo {author} {\bibfnamefont {S.}~\bibnamefont
  {Ren}}, \bibinfo {author} {\bibfnamefont {K.}~\bibnamefont {He}}, \bibinfo
  {author} {\bibfnamefont {R.}~\bibnamefont {Girshick}}, \ and\ \bibinfo
  {author} {\bibfnamefont {J.}~\bibnamefont {Sun}},\ }\href@noop {} {\bibfield
  {journal} {\bibinfo  {journal} {IEEE Trans. Pattern Anal. Mach. Intell.}\ }
  (\bibinfo {year} {2016})}\BibitemShut {NoStop}%
\bibitem [{\citenamefont {von Frisch}(1967)}]{Von_Frisch1967-bn}%
  \BibitemOpen
  \bibfield  {author} {\bibinfo {author} {\bibfnamefont {K.}~\bibnamefont {von
  Frisch}},\ }\href@noop {} {\emph {\bibinfo {title} {The Dance Language and
  Orientation of Bees}}},\ Vol.\ \bibinfo {volume} {159}\ (\bibinfo
  {publisher} {Belknap Press of Harvard University Press},\ \bibinfo {year}
  {1967})\BibitemShut {NoStop}%
\bibitem [{\citenamefont {Seeley}(2010)}]{Seeley2010-xn}%
  \BibitemOpen
  \bibfield  {author} {\bibinfo {author} {\bibfnamefont {T.~D.}\ \bibnamefont
  {Seeley}},\ }\href@noop {} {{\selectlanguage {english}\emph {\bibinfo {title}
  {Honeybee Democracy}}}}\ (\bibinfo  {publisher} {Princeton University
  Press},\ \bibinfo {year} {2010})\BibitemShut {NoStop}%
\bibitem [{\citenamefont {Winston}(1991)}]{Winston1991-vd}%
  \BibitemOpen
  \bibfield  {author} {\bibinfo {author} {\bibfnamefont {M.~L.}\ \bibnamefont
  {Winston}},\ }\href@noop {} {{\selectlanguage {english}\emph {\bibinfo
  {title} {The Biology of the Honey Bee}}}}\ (\bibinfo  {publisher} {Harvard
  University Press},\ \bibinfo {year} {1991})\BibitemShut {NoStop}%
\bibitem [{\citenamefont {Karaboga}\ and\ \citenamefont
  {Akay}(2009)}]{Karaboga2009-pa}%
  \BibitemOpen
  \bibfield  {author} {\bibinfo {author} {\bibfnamefont {D.}~\bibnamefont
  {Karaboga}}\ and\ \bibinfo {author} {\bibfnamefont {B.}~\bibnamefont
  {Akay}},\ }\href@noop {} {\bibfield  {journal} {\bibinfo  {journal} {Artif.
  Intell. Rev.}\ }\textbf {\bibinfo {volume} {31}},\ \bibinfo {pages} {61}
  (\bibinfo {year} {2009})}\BibitemShut {NoStop}%
\bibitem [{\citenamefont {Florea}(2013)}]{Florea2013-iu}%
  \BibitemOpen
  \bibfield  {author} {\bibinfo {author} {\bibfnamefont {M.~I.}\ \bibnamefont
  {Florea}},\ }{\selectlanguage {english}\emph {\bibinfo {title} {Automatic
  detection of honeybees in a hive}}},\ \href@noop {} {Ph.D. thesis} (\bibinfo
  {year} {2013})\BibitemShut {NoStop}%
\bibitem [{\citenamefont {Hendriks}\ \emph {et~al.}(2012)\citenamefont
  {Hendriks}, \citenamefont {Yu}, \citenamefont {Lecocq}, \citenamefont
  {Bakker}, \citenamefont {Locke},\ and\ \citenamefont
  {Terenius}}]{Hendriks2012-en}%
  \BibitemOpen
  \bibfield  {author} {\bibinfo {author} {\bibfnamefont {C.}~\bibnamefont
  {Hendriks}}, \bibinfo {author} {\bibfnamefont {Z.}~\bibnamefont {Yu}},
  \bibinfo {author} {\bibfnamefont {A.}~\bibnamefont {Lecocq}}, \bibinfo
  {author} {\bibfnamefont {T.}~\bibnamefont {Bakker}}, \bibinfo {author}
  {\bibfnamefont {B.}~\bibnamefont {Locke}}, \ and\ \bibinfo {author}
  {\bibfnamefont {O.}~\bibnamefont {Terenius}},\ }in\ \href@noop {} {\emph
  {\bibinfo {booktitle} {Workshop Vis. Observation Anal. Anim. Insect Behav.
  {ICPR}}}}\ (\bibinfo  {publisher} {Citeseer},\ \bibinfo {year}
  {2012})\BibitemShut {NoStop}%
\bibitem [{\citenamefont {Kimura}\ \emph {et~al.}(2011)\citenamefont {Kimura},
  \citenamefont {Ohashi}, \citenamefont {Okada},\ and\ \citenamefont
  {Ikeno}}]{Kimura2011-oy}%
  \BibitemOpen
  \bibfield  {author} {\bibinfo {author} {\bibfnamefont {T.}~\bibnamefont
  {Kimura}}, \bibinfo {author} {\bibfnamefont {M.}~\bibnamefont {Ohashi}},
  \bibinfo {author} {\bibfnamefont {R.}~\bibnamefont {Okada}}, \ and\ \bibinfo
  {author} {\bibfnamefont {H.}~\bibnamefont {Ikeno}},\ }\href@noop {}
  {\bibfield  {journal} {\bibinfo  {journal} {Apidologie}\ }\textbf {\bibinfo
  {volume} {42}},\ \bibinfo {pages} {607} (\bibinfo {year} {2011})}\BibitemShut
  {NoStop}%
\bibitem [{\citenamefont {Kimura}\ \emph {et~al.}(2014)\citenamefont {Kimura},
  \citenamefont {Ohashi}, \citenamefont {Crailsheim}, \citenamefont {Schmickl},
  \citenamefont {Okada}, \citenamefont {Radspieler},\ and\ \citenamefont
  {Ikeno}}]{Kimura2014-em}%
  \BibitemOpen
  \bibfield  {author} {\bibinfo {author} {\bibfnamefont {T.}~\bibnamefont
  {Kimura}}, \bibinfo {author} {\bibfnamefont {M.}~\bibnamefont {Ohashi}},
  \bibinfo {author} {\bibfnamefont {K.}~\bibnamefont {Crailsheim}}, \bibinfo
  {author} {\bibfnamefont {T.}~\bibnamefont {Schmickl}}, \bibinfo {author}
  {\bibfnamefont {R.}~\bibnamefont {Okada}}, \bibinfo {author} {\bibfnamefont
  {G.}~\bibnamefont {Radspieler}}, \ and\ \bibinfo {author} {\bibfnamefont
  {H.}~\bibnamefont {Ikeno}},\ }\href@noop {} {\bibfield  {journal} {\bibinfo
  {journal} {PLoS One}\ }\textbf {\bibinfo {volume} {9}},\ \bibinfo {pages}
  {e84656} (\bibinfo {year} {2014})}\BibitemShut {NoStop}%
\bibitem [{\citenamefont {Wario}\ \emph {et~al.}(2015)\citenamefont {Wario},
  \citenamefont {Wild}, \citenamefont {Couvillon}, \citenamefont {Rojas},\ and\
  \citenamefont {Landgraf}}]{Wario2015-fm}%
  \BibitemOpen
  \bibfield  {author} {\bibinfo {author} {\bibfnamefont {F.}~\bibnamefont
  {Wario}}, \bibinfo {author} {\bibfnamefont {B.}~\bibnamefont {Wild}},
  \bibinfo {author} {\bibfnamefont {M.~J.}\ \bibnamefont {Couvillon}}, \bibinfo
  {author} {\bibfnamefont {R.}~\bibnamefont {Rojas}}, \ and\ \bibinfo {author}
  {\bibfnamefont {T.}~\bibnamefont {Landgraf}},\ }\href@noop {} {\bibfield
  {journal} {\bibinfo  {journal} {Front. Ecol. Evol.}\ }\textbf {\bibinfo
  {volume} {3}} (\bibinfo {year} {2015})}\BibitemShut {NoStop}%
\bibitem [{\citenamefont {Biesmeijer}\ and\ \citenamefont
  {Seeley}(2005)}]{Biesmeijer2005-dr}%
  \BibitemOpen
  \bibfield  {author} {\bibinfo {author} {\bibfnamefont {J.~C.}\ \bibnamefont
  {Biesmeijer}}\ and\ \bibinfo {author} {\bibfnamefont {T.~D.}\ \bibnamefont
  {Seeley}},\ }\href@noop {} {\bibfield  {journal} {\bibinfo  {journal} {Behav.
  Ecol. Sociobiol.}\ }\textbf {\bibinfo {volume} {59}},\ \bibinfo {pages} {133}
  (\bibinfo {year} {2005})}\BibitemShut {NoStop}%
\bibitem [{\citenamefont {Mersch}\ \emph {et~al.}(2013)\citenamefont {Mersch},
  \citenamefont {Crespi},\ and\ \citenamefont {Keller}}]{Mersch2013-uf}%
  \BibitemOpen
  \bibfield  {author} {\bibinfo {author} {\bibfnamefont {D.~P.}\ \bibnamefont
  {Mersch}}, \bibinfo {author} {\bibfnamefont {A.}~\bibnamefont {Crespi}}, \
  and\ \bibinfo {author} {\bibfnamefont {L.}~\bibnamefont {Keller}},\
  }\href@noop {} {\bibfield  {journal} {\bibinfo  {journal} {Science}\ }\textbf
  {\bibinfo {volume} {340}},\ \bibinfo {pages} {1090} (\bibinfo {year}
  {2013})}\BibitemShut {NoStop}%
\bibitem [{\citenamefont {Kratz}\ and\ \citenamefont
  {Nishino}(2010)}]{Kratz2010-kb}%
  \BibitemOpen
  \bibfield  {author} {\bibinfo {author} {\bibfnamefont {L.}~\bibnamefont
  {Kratz}}\ and\ \bibinfo {author} {\bibfnamefont {K.}~\bibnamefont
  {Nishino}},\ }in\ \href@noop {} {\emph {\bibinfo {booktitle} {Computer Vision
  and Pattern Recognition ({CVPR)}, 2010 {IEEE} Conference on}}}\ (\bibinfo
  {publisher} {ieeexplore.ieee.org},\ \bibinfo {year} {2010})\ pp.\ \bibinfo
  {pages} {693--700}\BibitemShut {NoStop}%
\bibitem [{\citenamefont {Ali}\ and\ \citenamefont
  {Dailey}(2009)}]{Ali2009-sf}%
  \BibitemOpen
  \bibfield  {author} {\bibinfo {author} {\bibfnamefont {I.}~\bibnamefont
  {Ali}}\ and\ \bibinfo {author} {\bibfnamefont {M.~N.}\ \bibnamefont
  {Dailey}},\ }in\ \href@noop {} {{\selectlanguage {english}\emph {\bibinfo
  {booktitle} {Advanced Concepts for Intelligent Vision Systems}}}},\ \bibinfo
  {series and number} {Lecture Notes in Computer Science},\ \bibinfo {editor}
  {edited by\ \bibinfo {editor} {\bibfnamefont {J.}~\bibnamefont
  {Blanc-Talon}}, \bibinfo {editor} {\bibfnamefont {W.}~\bibnamefont
  {Philips}}, \bibinfo {editor} {\bibfnamefont {D.}~\bibnamefont {Popescu}}, \
  and\ \bibinfo {editor} {\bibfnamefont {P.}~\bibnamefont {Scheunders}}}\
  (\bibinfo  {publisher} {Springer Berlin Heidelberg},\ \bibinfo {year}
  {2009})\ pp.\ \bibinfo {pages} {540--549}\BibitemShut {NoStop}%
\bibitem [{\citenamefont {Ge}\ \emph {et~al.}(2012)\citenamefont {Ge},
  \citenamefont {Collins},\ and\ \citenamefont {Ruback}}]{Ge2012-tz}%
  \BibitemOpen
  \bibfield  {author} {\bibinfo {author} {\bibfnamefont {W.}~\bibnamefont
  {Ge}}, \bibinfo {author} {\bibfnamefont {R.~T.}\ \bibnamefont {Collins}}, \
  and\ \bibinfo {author} {\bibfnamefont {R.~B.}\ \bibnamefont {Ruback}},\
  }\href@noop {} {\bibfield  {journal} {\bibinfo  {journal} {IEEE Trans.
  Pattern Anal. Mach. Intell.}\ }\textbf {\bibinfo {volume} {34}},\ \bibinfo
  {pages} {1003} (\bibinfo {year} {2012})}\BibitemShut {NoStop}%
\bibitem [{\citenamefont {Rodriguez}\ \emph {et~al.}(2011)\citenamefont
  {Rodriguez}, \citenamefont {Sivic}, \citenamefont {Laptev},\ and\
  \citenamefont {Audibert}}]{Rodriguez2011-qo}%
  \BibitemOpen
  \bibfield  {author} {\bibinfo {author} {\bibfnamefont {M.}~\bibnamefont
  {Rodriguez}}, \bibinfo {author} {\bibfnamefont {J.}~\bibnamefont {Sivic}},
  \bibinfo {author} {\bibfnamefont {I.}~\bibnamefont {Laptev}}, \ and\ \bibinfo
  {author} {\bibfnamefont {J.~Y.}\ \bibnamefont {Audibert}},\ }in\ \href@noop
  {} {\emph {\bibinfo {booktitle} {2011 International Conference on Computer
  Vision}}}\ (\bibinfo {year} {2011})\ pp.\ \bibinfo {pages}
  {1235--1242}\BibitemShut {NoStop}%
\bibitem [{\citenamefont {Girshick}(2015)}]{Girshick2015-tj}%
  \BibitemOpen
  \bibfield  {author} {\bibinfo {author} {\bibfnamefont {R.}~\bibnamefont
  {Girshick}},\ }in\ \href@noop {} {\emph {\bibinfo {booktitle} {Proceedings of
  the {IEEE} International Conference on Computer Vision}}}\ (\bibinfo
  {publisher} {cv-foundation.org},\ \bibinfo {year} {2015})\ pp.\ \bibinfo
  {pages} {1440--1448}\BibitemShut {NoStop}%
\bibitem [{\citenamefont {Hosang}\ \emph {et~al.}(2016)\citenamefont {Hosang},
  \citenamefont {Benenson}, \citenamefont {Doll{\'a}r},\ and\ \citenamefont
  {Schiele}}]{Hosang2016-lo}%
  \BibitemOpen
  \bibfield  {author} {\bibinfo {author} {\bibfnamefont {J.}~\bibnamefont
  {Hosang}}, \bibinfo {author} {\bibfnamefont {R.}~\bibnamefont {Benenson}},
  \bibinfo {author} {\bibfnamefont {P.}~\bibnamefont {Doll{\'a}r}}, \ and\
  \bibinfo {author} {\bibfnamefont {B.}~\bibnamefont {Schiele}},\ }\href@noop
  {} {\bibfield  {journal} {\bibinfo  {journal} {IEEE Trans. Pattern Anal.
  Mach. Intell.}\ }\textbf {\bibinfo {volume} {38}},\ \bibinfo {pages} {814}
  (\bibinfo {year} {2016})}\BibitemShut {NoStop}%
\bibitem [{\citenamefont {Pinheiro}\ \emph {et~al.}(2015)\citenamefont
  {Pinheiro}, \citenamefont {Collobert},\ and\ \citenamefont
  {Dollar}}]{Pinheiro2015-ht}%
  \BibitemOpen
  \bibfield  {author} {\bibinfo {author} {\bibfnamefont {P.~O.}\ \bibnamefont
  {Pinheiro}}, \bibinfo {author} {\bibfnamefont {R.}~\bibnamefont {Collobert}},
  \ and\ \bibinfo {author} {\bibfnamefont {P.}~\bibnamefont {Dollar}},\
  }\href@noop {} {\  (\bibinfo {year} {2015})},\ \Eprint
  {http://arxiv.org/abs/1506.06204} {arXiv:1506.06204 [cs.CV]} \BibitemShut
  {NoStop}%
\bibitem [{\citenamefont {Dai}\ \emph {et~al.}(2015)\citenamefont {Dai},
  \citenamefont {He},\ and\ \citenamefont {Sun}}]{Dai2015-tz}%
  \BibitemOpen
  \bibfield  {author} {\bibinfo {author} {\bibfnamefont {J.}~\bibnamefont
  {Dai}}, \bibinfo {author} {\bibfnamefont {K.}~\bibnamefont {He}}, \ and\
  \bibinfo {author} {\bibfnamefont {J.}~\bibnamefont {Sun}},\ }\href@noop {} {\
   (\bibinfo {year} {2015})},\ \Eprint {http://arxiv.org/abs/1512.04412}
  {arXiv:1512.04412 [cs.CV]} \BibitemShut {NoStop}%
\bibitem [{\citenamefont {Ondruska}\ and\ \citenamefont
  {Posner}(2016)}]{Ondruska2016-dd}%
  \BibitemOpen
  \bibfield  {author} {\bibinfo {author} {\bibfnamefont {P.}~\bibnamefont
  {Ondruska}}\ and\ \bibinfo {author} {\bibfnamefont {I.}~\bibnamefont
  {Posner}},\ }\href@noop {} {\  (\bibinfo {year} {2016})},\ \Eprint
  {http://arxiv.org/abs/1602.00991} {arXiv:1602.00991 [cs.LG]} \BibitemShut
  {NoStop}%
\bibitem [{\citenamefont {Chen}(2003)}]{Chen2003-er}%
  \BibitemOpen
  \bibfield  {author} {\bibinfo {author} {\bibfnamefont {Z.}~\bibnamefont
  {Chen}},\ }\href@noop {} {\bibfield  {journal} {\bibinfo  {journal}
  {Statistics}\ }\textbf {\bibinfo {volume} {182}},\ \bibinfo {pages} {1}
  (\bibinfo {year} {2003})}\BibitemShut {NoStop}%
\bibitem [{\citenamefont {Ronneberger}\ \emph {et~al.}(2015)\citenamefont
  {Ronneberger}, \citenamefont {Fischer},\ and\ \citenamefont
  {Brox}}]{Ronneberger2015-bx}%
  \BibitemOpen
  \bibfield  {author} {\bibinfo {author} {\bibfnamefont {O.}~\bibnamefont
  {Ronneberger}}, \bibinfo {author} {\bibfnamefont {P.}~\bibnamefont
  {Fischer}}, \ and\ \bibinfo {author} {\bibfnamefont {T.}~\bibnamefont
  {Brox}},\ }in\ \href@noop {} {{\selectlanguage {english}\emph {\bibinfo
  {booktitle} {Medical Image Computing and {Computer-Assisted} Intervention --
  {MICCAI} 2015}}}},\ \bibinfo {series and number} {Lecture Notes in Computer
  Science},\ \bibinfo {editor} {edited by\ \bibinfo {editor} {\bibfnamefont
  {N.}~\bibnamefont {Navab}}, \bibinfo {editor} {\bibfnamefont
  {J.}~\bibnamefont {Hornegger}}, \bibinfo {editor} {\bibfnamefont {W.~M.}\
  \bibnamefont {Wells}}, \ and\ \bibinfo {editor} {\bibfnamefont {A.~F.}\
  \bibnamefont {Frangi}}}\ (\bibinfo  {publisher} {Springer International
  Publishing},\ \bibinfo {year} {2015})\ pp.\ \bibinfo {pages}
  {234--241}\BibitemShut {NoStop}%
\bibitem [{\citenamefont {Kingma}\ and\ \citenamefont
  {Ba}(2014)}]{Kingma2014-ln}%
  \BibitemOpen
  \bibfield  {author} {\bibinfo {author} {\bibfnamefont {D.}~\bibnamefont
  {Kingma}}\ and\ \bibinfo {author} {\bibfnamefont {J.}~\bibnamefont {Ba}},\
  }\href@noop {} {\  (\bibinfo {year} {2014})},\ \Eprint
  {http://arxiv.org/abs/1412.6980} {arXiv:1412.6980 [cs.LG]} \BibitemShut
  {NoStop}%
\bibitem [{\citenamefont {Srivastava}\ \emph {et~al.}()\citenamefont
  {Srivastava}, \citenamefont {Hinton}, \citenamefont {Krizhevsky},
  \citenamefont {Sutskever},\ and\ \citenamefont
  {Salakhutdinov}}]{Srivastava_undated-eq}%
  \BibitemOpen
  \bibfield  {author} {\bibinfo {author} {\bibfnamefont {N.}~\bibnamefont
  {Srivastava}}, \bibinfo {author} {\bibfnamefont {G.~R.}\ \bibnamefont
  {Hinton}}, \bibinfo {author} {\bibfnamefont {A.}~\bibnamefont {Krizhevsky}},
  \bibinfo {author} {\bibfnamefont {I.}~\bibnamefont {Sutskever}}, \ and\
  \bibinfo {author} {\bibfnamefont {R.}~\bibnamefont {Salakhutdinov}},\
  }\href@noop {} {\ }\BibitemShut {NoStop}%
\bibitem [{\citenamefont {Park}\ and\ \citenamefont
  {Kwak}()}]{Park_undated-jg}%
  \BibitemOpen
  \bibfield  {author} {\bibinfo {author} {\bibfnamefont {S.}~\bibnamefont
  {Park}}\ and\ \bibinfo {author} {\bibfnamefont {N.}~\bibnamefont {Kwak}},\
  }\href@noop {} {\ }\BibitemShut {NoStop}%
\bibitem [{\citenamefont {Springenberg}\ \emph {et~al.}(2014)\citenamefont
  {Springenberg}, \citenamefont {Dosovitskiy}, \citenamefont {Brox},\ and\
  \citenamefont {Riedmiller}}]{Springenberg2014-xu}%
  \BibitemOpen
  \bibfield  {author} {\bibinfo {author} {\bibfnamefont {J.~T.}\ \bibnamefont
  {Springenberg}}, \bibinfo {author} {\bibfnamefont {A.}~\bibnamefont
  {Dosovitskiy}}, \bibinfo {author} {\bibfnamefont {T.}~\bibnamefont {Brox}}, \
  and\ \bibinfo {author} {\bibfnamefont {M.}~\bibnamefont {Riedmiller}},\
  }\href@noop {} {\  (\bibinfo {year} {2014})},\ \Eprint
  {http://arxiv.org/abs/1412.6806} {arXiv:1412.6806 [cs.LG]} \BibitemShut
  {NoStop}%
\bibitem [{\citenamefont {Dieleman}\ \emph {et~al.}(2015)\citenamefont
  {Dieleman}, \citenamefont {Willett},\ and\ \citenamefont
  {Dambre}}]{Dieleman2015-zi}%
  \BibitemOpen
  \bibfield  {author} {\bibinfo {author} {\bibfnamefont {S.}~\bibnamefont
  {Dieleman}}, \bibinfo {author} {\bibfnamefont {K.~W.}\ \bibnamefont
  {Willett}}, \ and\ \bibinfo {author} {\bibfnamefont {J.}~\bibnamefont
  {Dambre}},\ }\href@noop {} {\  (\bibinfo {year} {2015})},\ \Eprint
  {http://arxiv.org/abs/1503.07077} {arXiv:1503.07077 [astro-ph.IM]}
  \BibitemShut {NoStop}%
\bibitem [{\citenamefont {Marcos}\ \emph {et~al.}(2016)\citenamefont {Marcos},
  \citenamefont {Volpi},\ and\ \citenamefont {Tuia}}]{Marcos2016-es}%
  \BibitemOpen
  \bibfield  {author} {\bibinfo {author} {\bibfnamefont {D.}~\bibnamefont
  {Marcos}}, \bibinfo {author} {\bibfnamefont {M.}~\bibnamefont {Volpi}}, \
  and\ \bibinfo {author} {\bibfnamefont {D.}~\bibnamefont {Tuia}},\ }\href@noop
  {} {\  (\bibinfo {year} {2016})},\ \Eprint {http://arxiv.org/abs/1604.06720}
  {arXiv:1604.06720 [cs.CV]} \BibitemShut {NoStop}%
\bibitem [{\citenamefont {Worrall}\ \emph {et~al.}(2016)\citenamefont
  {Worrall}, \citenamefont {Garbin}, \citenamefont {Turmukhambetov},\ and\
  \citenamefont {Brostow}}]{Worrall2016-up}%
  \BibitemOpen
  \bibfield  {author} {\bibinfo {author} {\bibfnamefont {D.~E.}\ \bibnamefont
  {Worrall}}, \bibinfo {author} {\bibfnamefont {S.~J.}\ \bibnamefont {Garbin}},
  \bibinfo {author} {\bibfnamefont {D.}~\bibnamefont {Turmukhambetov}}, \ and\
  \bibinfo {author} {\bibfnamefont {G.~J.}\ \bibnamefont {Brostow}},\
  }\href@noop {} {\  (\bibinfo {year} {2016})},\ \Eprint
  {http://arxiv.org/abs/1612.04642} {arXiv:1612.04642 [cs.CV]} \BibitemShut
  {NoStop}%
\end{thebibliography}%
\clearpage
\onecolumngrid

\section*{Supplementary Material}

\setcounter{figure}{0} 
\renewcommand*{\thefigure}{S\arabic{figure}}

\begin{figure}[htb]
  \centering
  \includegraphics[width=0.9\columnwidth]{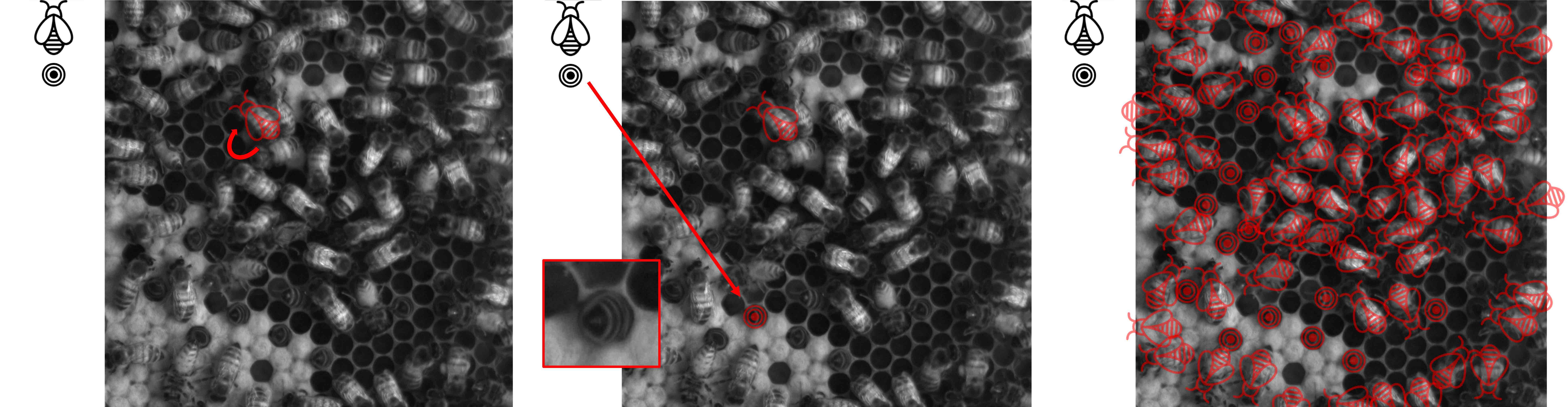}
  \caption{Amazon Mechanical Turk labelling schematic. Annotators were instructed to mark all the bees in an image of a bee comb through dragging and dropping a bee symbol on each bee in the image and matching the symbol's orientation angle. There are 2 symbols for marking - a bee symbol to mark fully visible bees and circle symbol to mark bees that are inside the cells, where only the bee abdomen is visible. Annotators are also instructed to use the same symbols to mark the small number of bees that are upside down.}
  \label{fig:AMT-schematic}
\end{figure}
\bigskip
\bigskip

\begin{figure}[htb]
  \centering
  \includegraphics[width=0.6\columnwidth]{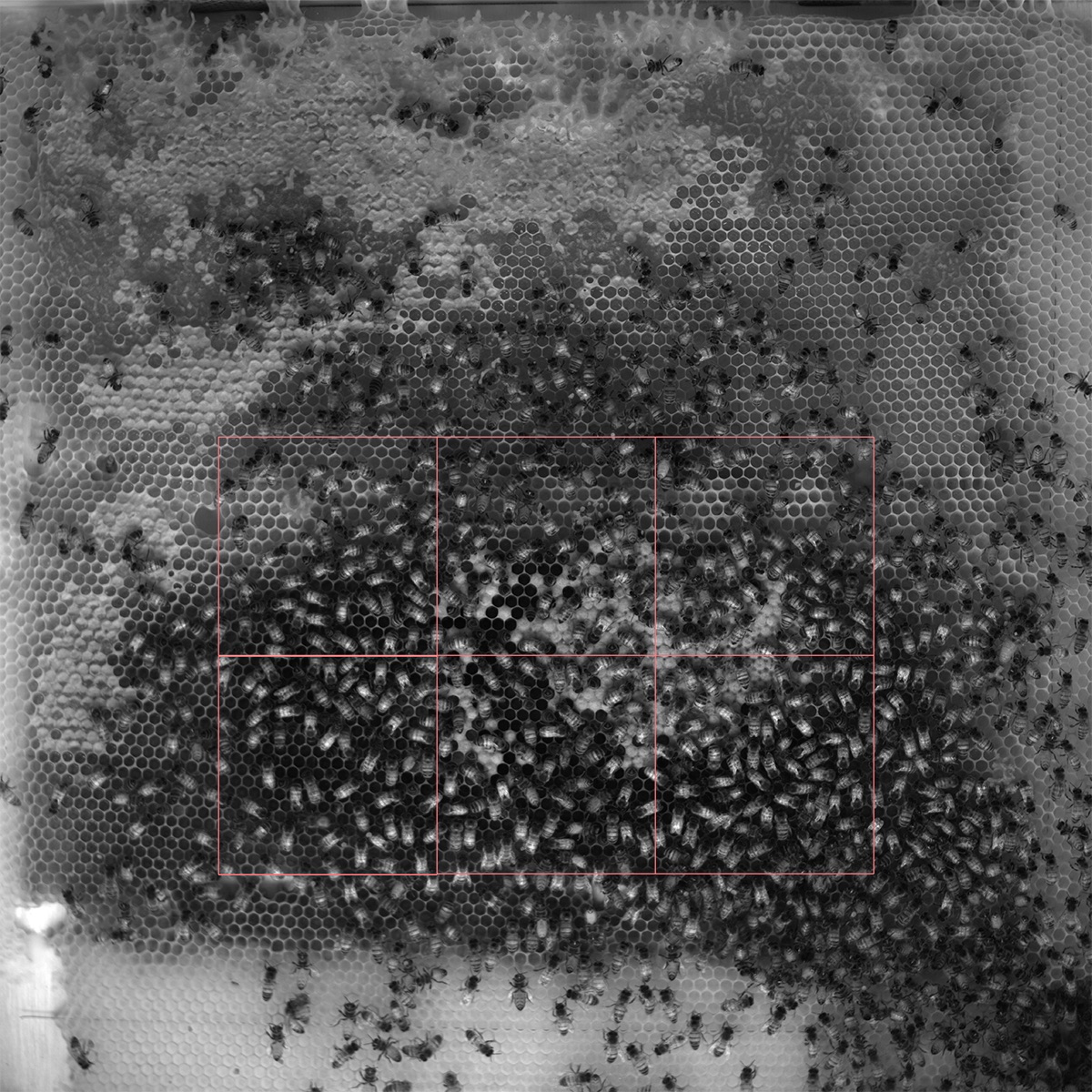}
  \caption{Regions of the 30 FPS beehive video used for human labeling and network training. The squares designate the size of subregions used as one Amazon Mechanical Turk task.}
  \label{fig:marked_frames30}
\end{figure}

\begin{figure}[htb]
  \centering
  \includegraphics[width=0.6\columnwidth]{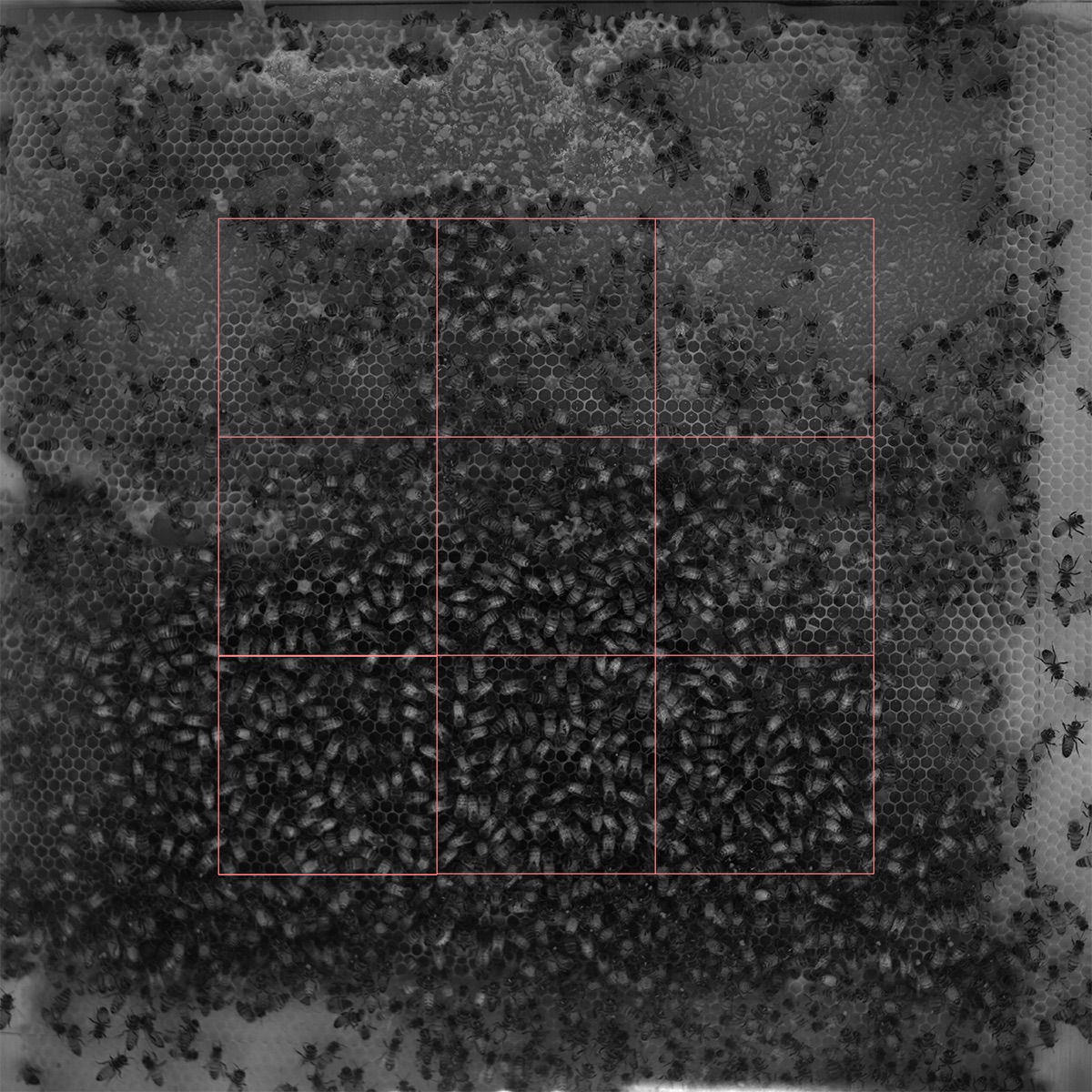}
  \caption{ Regions of the 70 FPS beehive video used for human labeling and network training. The squares designate the size of subregions used as one Amazon Mechanical Turk task.}\label{fig:marked_frames70}
\end{figure}

\begin{figure}[h]
  \centering
  \includegraphics[width=0.9\columnwidth]{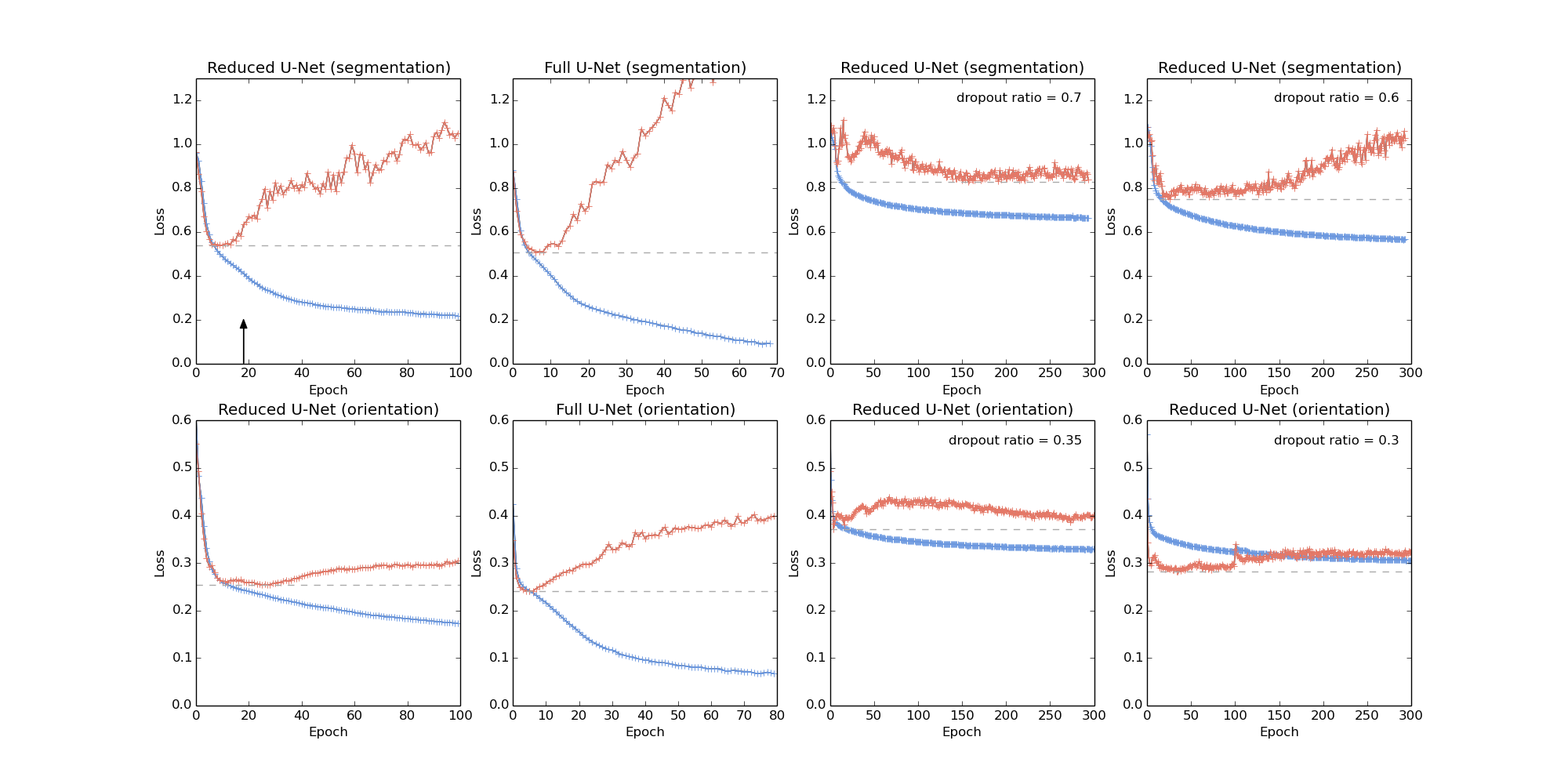}
  \caption{ The change of the loss function with training epoch using reduced U-Net (first column), full U-Net (second column) and different levels of dropout (column 3 and 4). The network performing the segmentation task is shown in the first row, the network performing orientation angle search in the second row. The loss for the training set is marked in blue and the loss for the test set in red. The full U-Net results in substantial overfitting, while reducing the size of U-Net reduces the amount of overfitting. Various levels of dropout result in prohibitively slow training (3rd column), or also lead to overfitting with worse overall results (4th column). For this reason we chose early-stop  in training (iteration 18 indicated on the upper-left panel) as a measure against overfitting.}
 \label{fig:dropout}
\end{figure}

\begin{figure}[htb]
  \centering
  \includegraphics[width=0.65\textwidth]{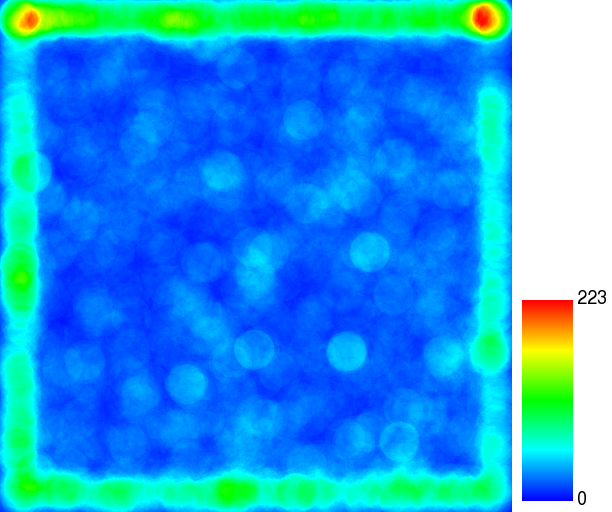}
  \caption{ Edge effects of the input image reduce network performance. We show a spatial histogram of the number of incorrectly predicted bees (FP's) across the 512x512 pixel image patches used as input to the network.}
 \label{fig:unet_hm}
\end{figure}

\begin{figure}[htb]
  \centering
    \begin{subfigure}[b]{\textwidth}
    \centering
        \includegraphics[width=0.65\textwidth]{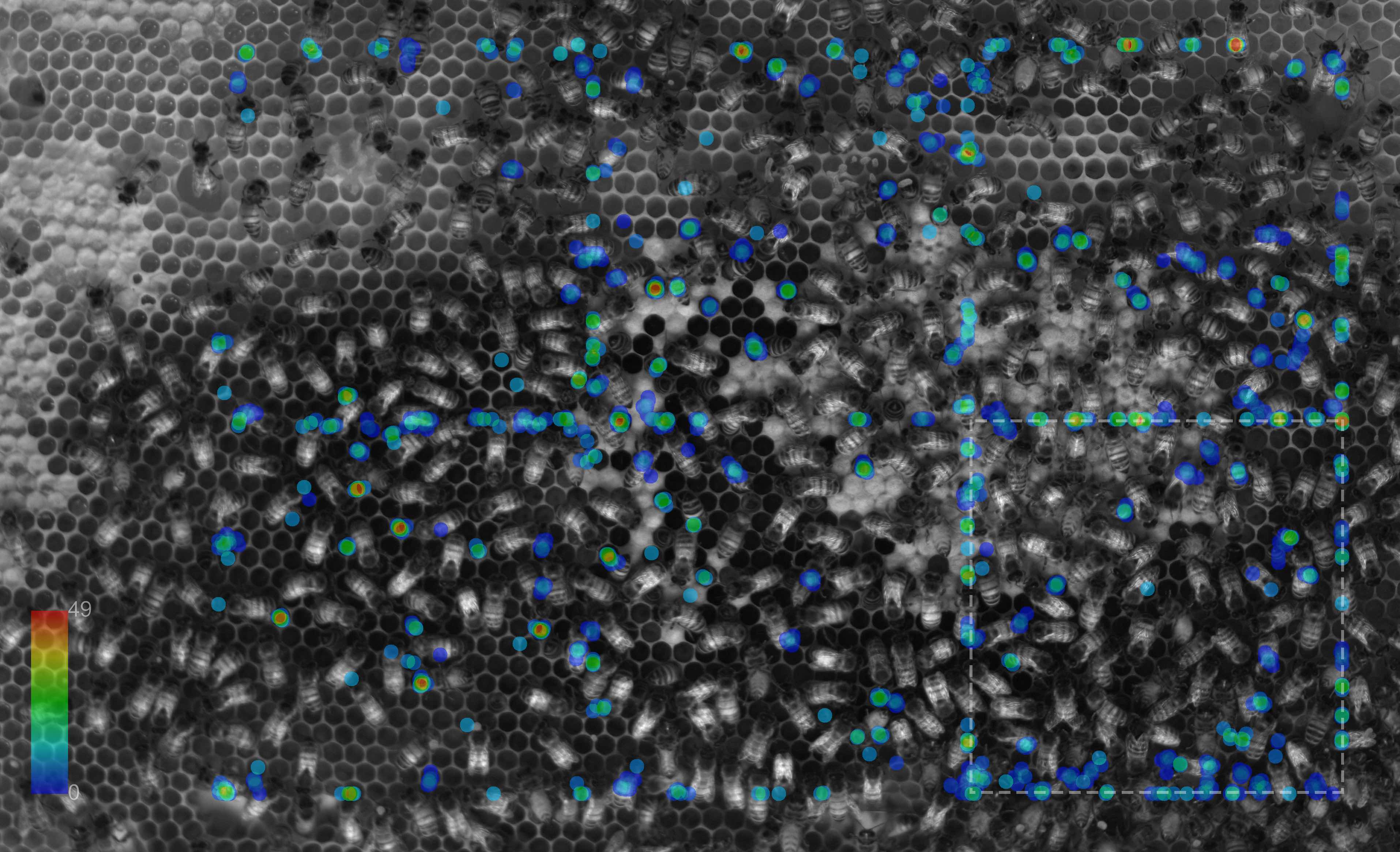}
        \caption{30\,FPS recording}
    \end{subfigure}
    \\
        \begin{subfigure}[b]{\textwidth}
        \centering
        \includegraphics[width=0.65\textwidth]{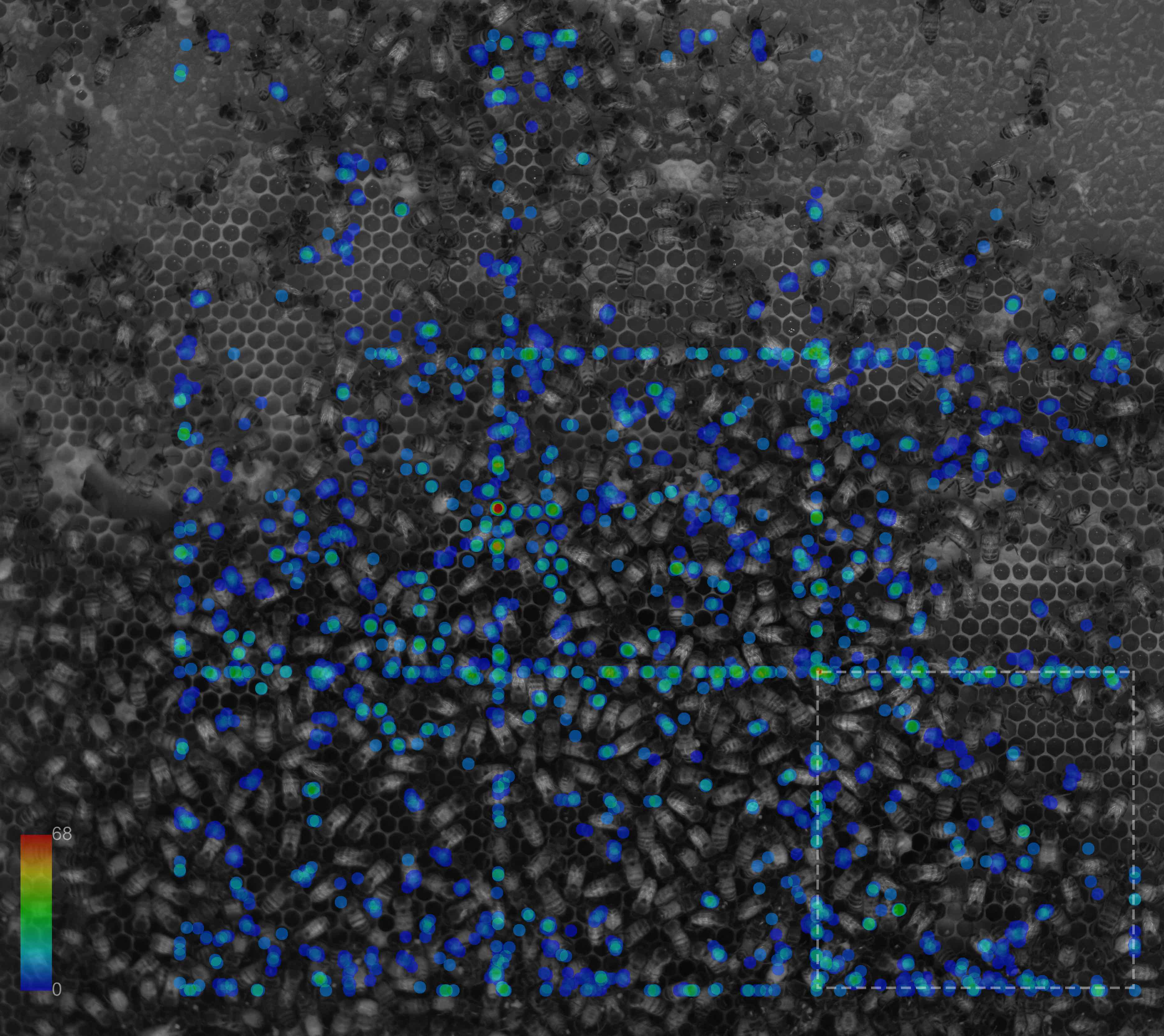}
        \caption{70\,FPS recording}
    \end{subfigure}
\caption{ Edge effects of the input image increase variability among human annotators.  We show a spatial histogram of the location of disagreements in bee labeling among human raters. In an individual AMT task,  annotators labeled 1024x1024 pixel image patch, one of which is indicated with a square shape in the images (white dashed outline). }
\label{fig:human_frame}
\end{figure}

\begin{figure}[h]
  \centering
  \includegraphics[width=0.5\textwidth]{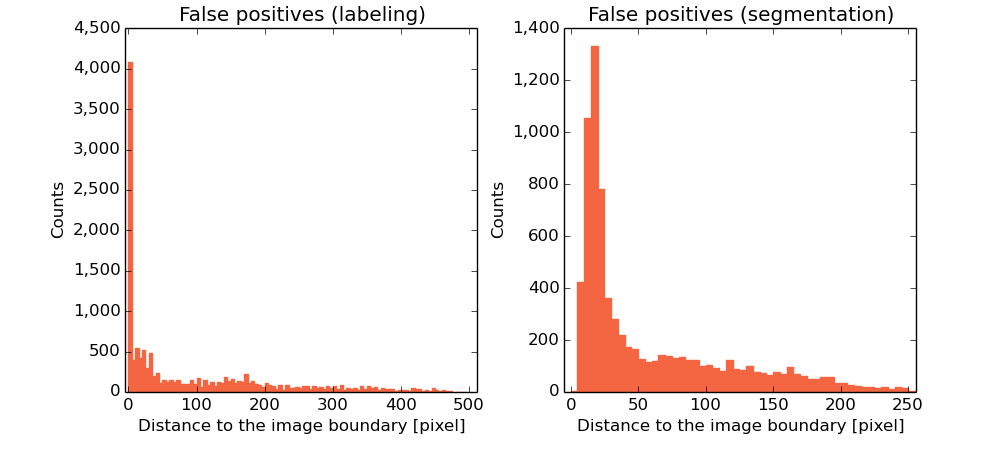}
  \caption{ The number of the labeling disagreements among AMT annotators (left) and FP's as identified through the network segmentation (right) is large near the image patch boundary suggesting an edge effect that can be improved in later implementations.}\label{fig:boundary_vs_err}
\end{figure}

\begin{figure}[h]
  \centering
        \includegraphics[width=0.20\textwidth]{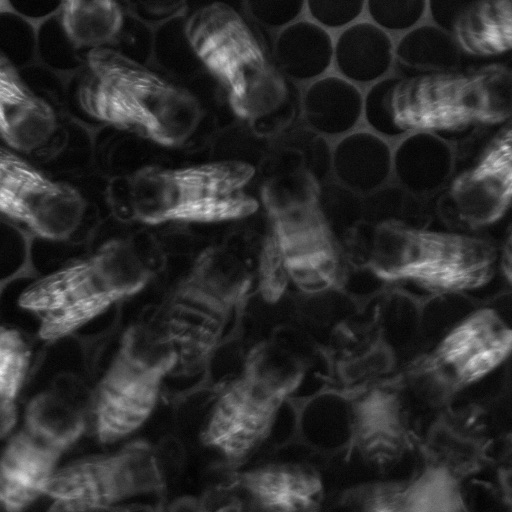}
        \includegraphics[width=0.20\textwidth]{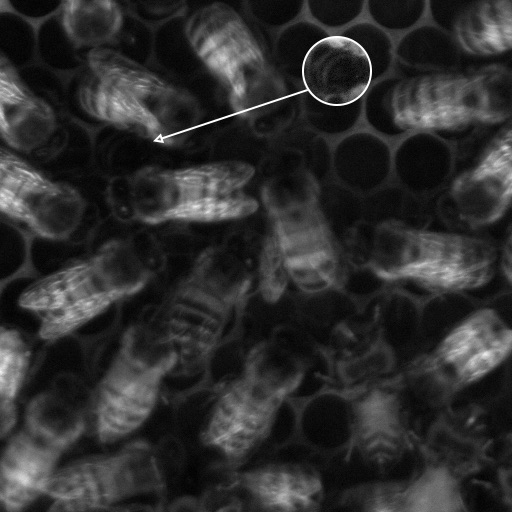}
        \includegraphics[width=0.20\textwidth]{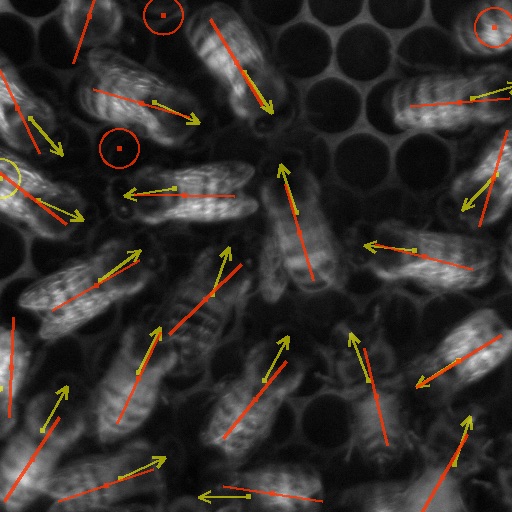}
    \\
        \includegraphics[width=0.20\textwidth]{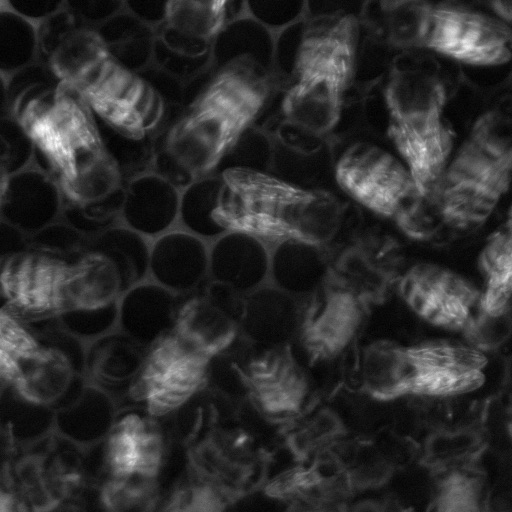}
        \includegraphics[width=0.20\textwidth]{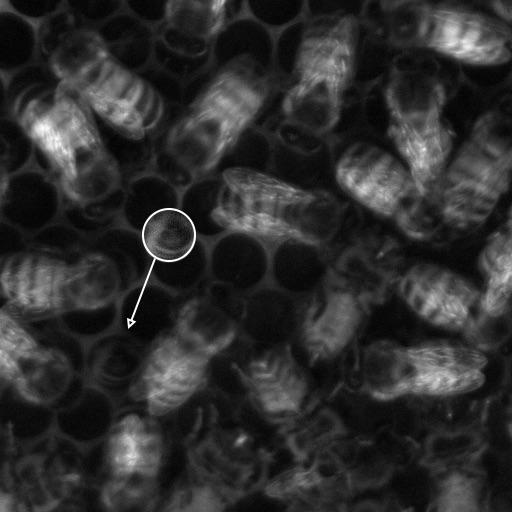}
        \includegraphics[width=0.20\textwidth]{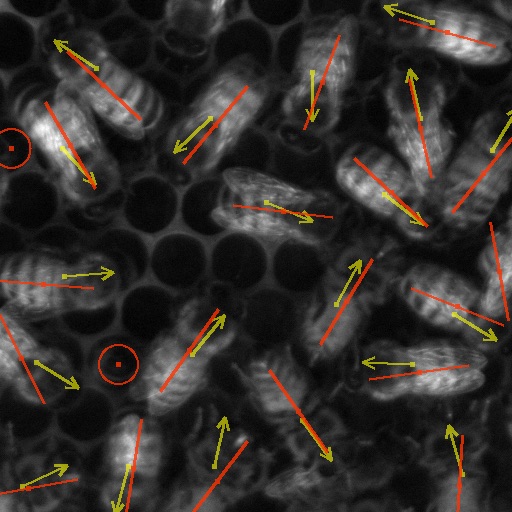}
      \\
        \includegraphics[width=0.20\textwidth]{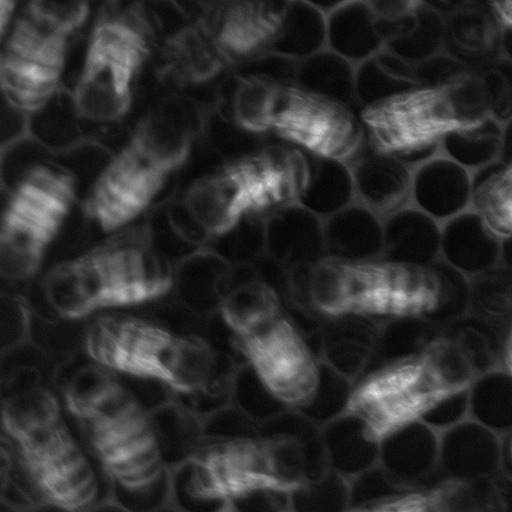}
        \includegraphics[width=0.20\textwidth]{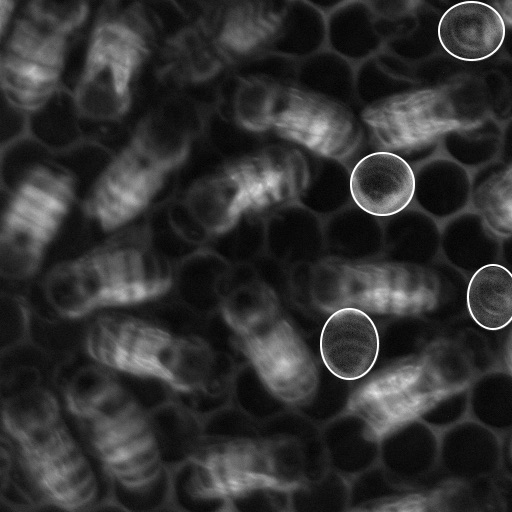}
        \includegraphics[width=0.20\textwidth]{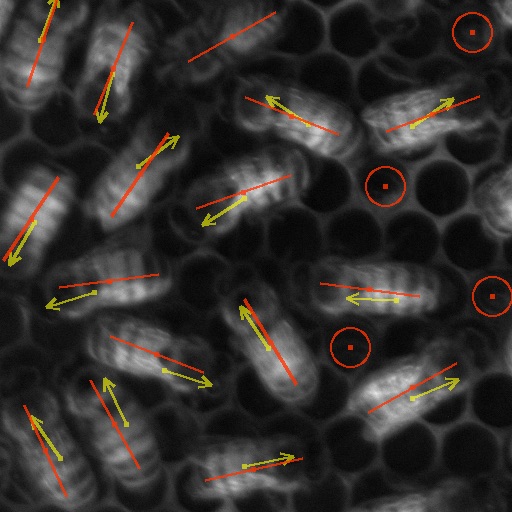}
      \\
        \includegraphics[width=0.20\textwidth]{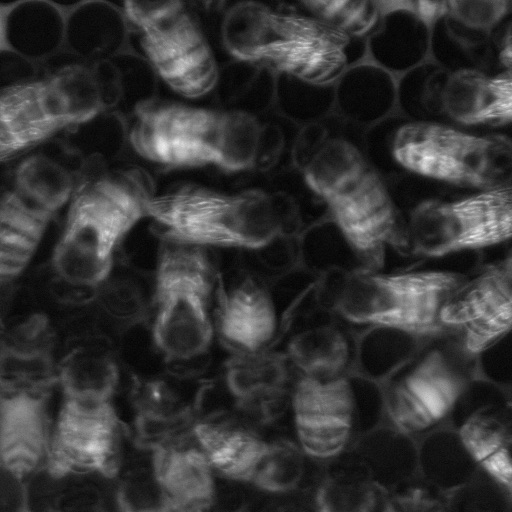}
        \includegraphics[width=0.20\textwidth]{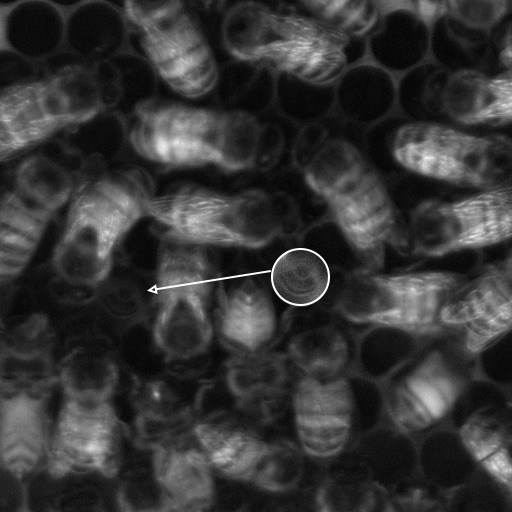}
        \includegraphics[width=0.20\textwidth]{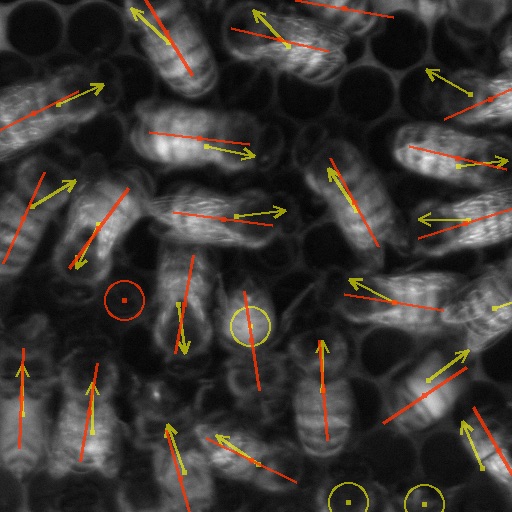}
    \caption{Bees which are partially obscured inside comb cells are often hard to identify by human labelers but still correctly segmented.  Shown are examples of difficult to label cases. For the original image (left column), the bee abdomens unnoticed by labelers are highlighted (middle column). These tails were however picked by the segmentation network, which is shown in the corresponding images in the right panel with labels marked in yellow and predictions in red.  Such cases contribute to number of FP's in the network performance reported in Table I.}\label{fig:butts}
\end{figure}

\begin{figure}[htb]
  \centering
  \includegraphics[width=0.7\columnwidth]{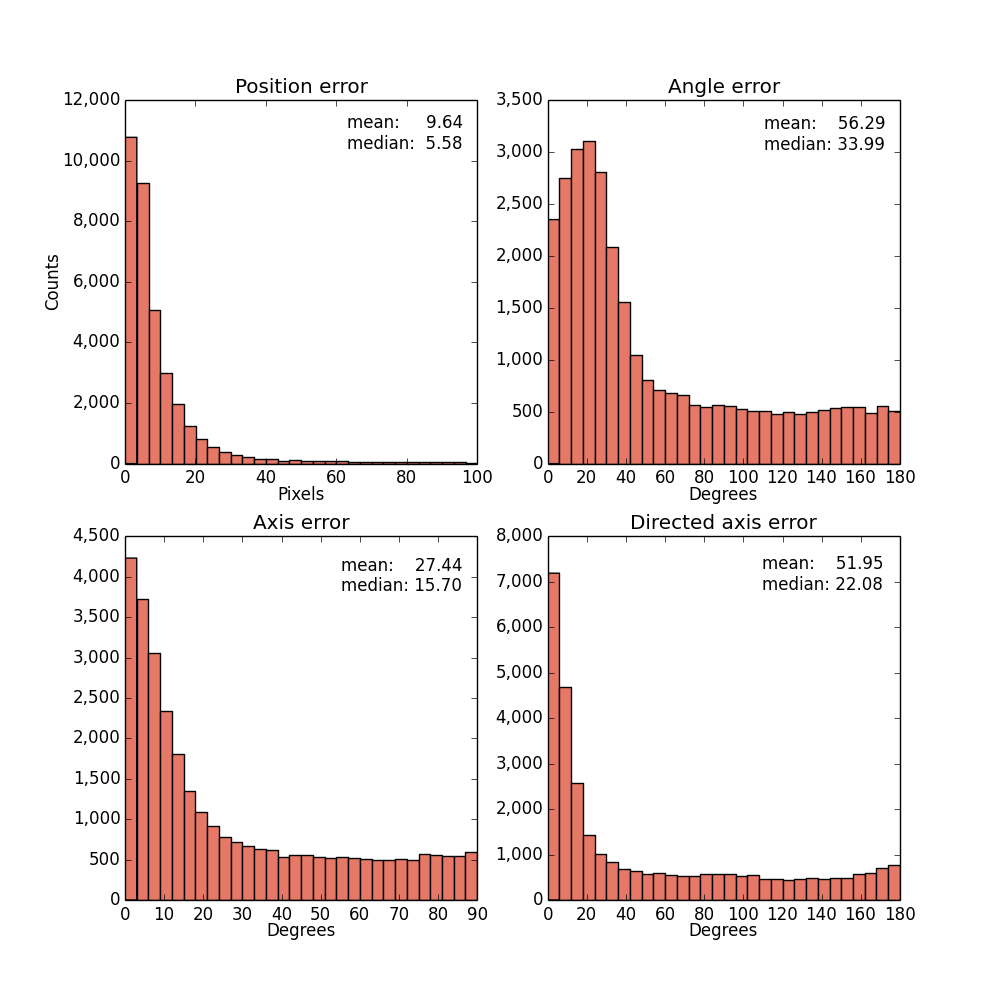}
  \caption{Network prediction errors for all labelled bee instances in the 2,176 images of the test dataset. We show error histograms as well as the mean and median errors for position, orientation angle, axis angle and directed axis angle predictions. The flat tails for angle predictions suggest a small baseline of random predictions. }
\label{fig:error_tail}
\end{figure}

\begin{figure}[htb]
  \centering
  \includegraphics[width=0.8\columnwidth]{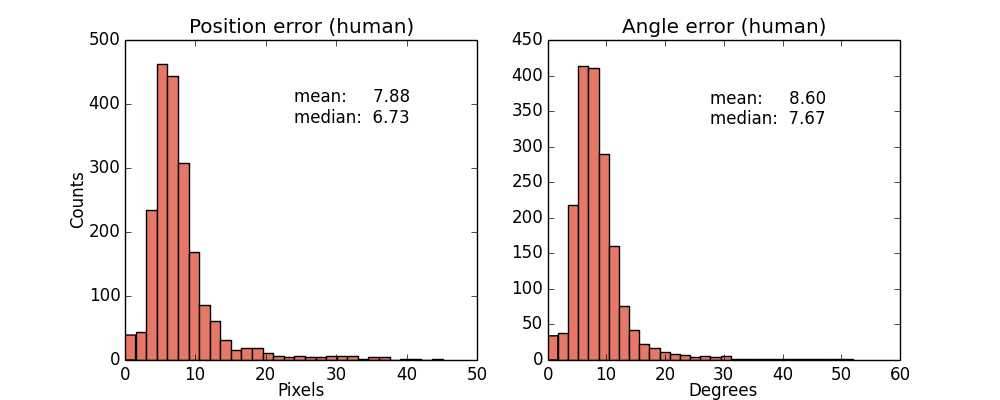}
  \caption{ Variability in annotated bee position and orientation among human raters. For 2034 bee instances we show the histogram of the standard deviation of 10 repeated annotation tasks against the one reference annotation used in the dataset for network training and testing.}\label{fig:human_distribution}
\end{figure}

\begin{figure}[h]
  \centering
  \includegraphics[width=0.7\columnwidth]{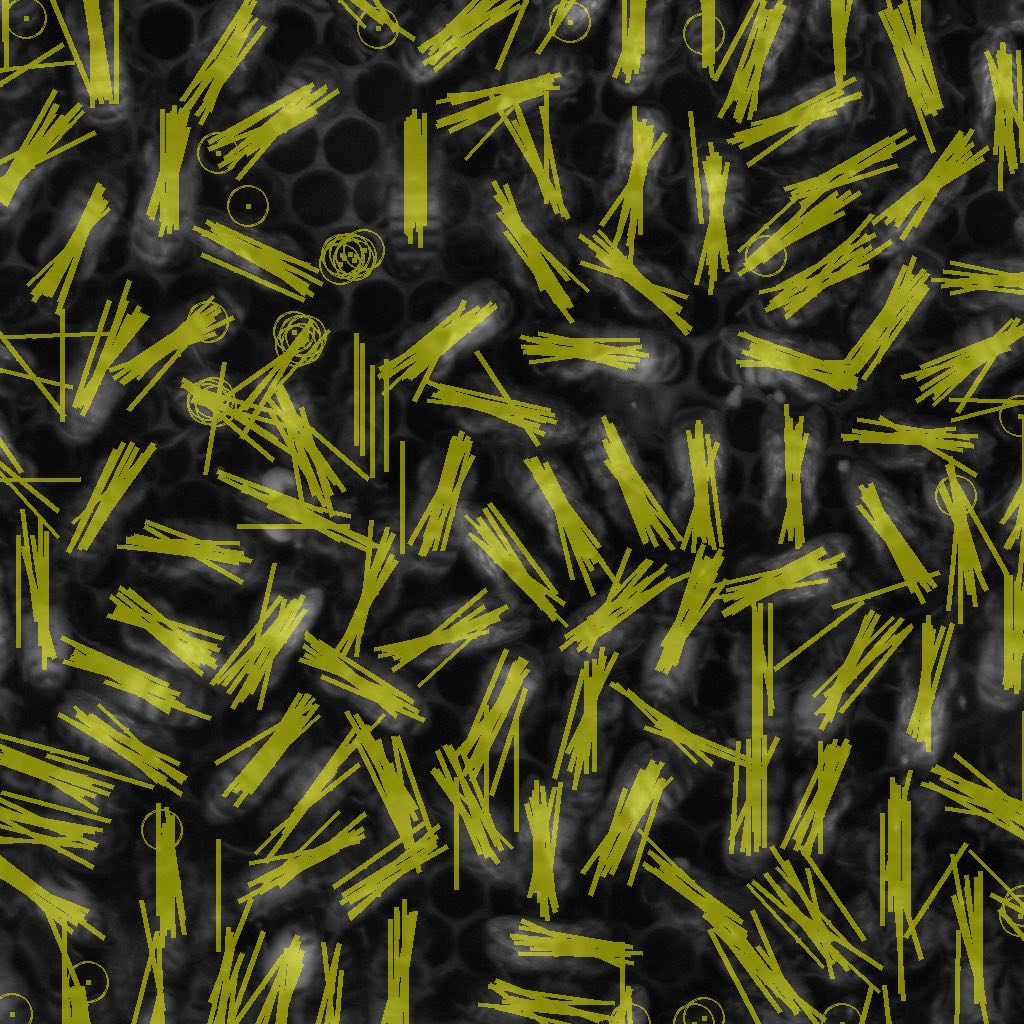}
  \caption{Example of the variability among human raters. Each yellow line is centered and aligned to a honeybee body identified by an annotator from Amazon Mechanical Turk and the same image was presented across 10 different annotators.  Circles are centered on image locations identified as bee abdomens.}\label{fig:human_example}
\end{figure}

\begin{figure}[htb]
\centering
\caption*{{\bf Supplementary Movie:} In the ancillary file ``simple\_tracking.mp4'' we show an example of a reconstructed trajectories.  Individuals in one frame are matched to the closest individuals in following frames using position, orientation, angle, and velocity.  In case a trajectory is lost, we searched up to five frames ahead for a close match to complete this trajectory.}
\end{figure}

\end{document}